\documentclass{article}
\usepackage{iclr2027_conference,times}

\usepackage{amsmath,amsfonts,bm}

\def\eqref#1{equation~\ref{#1}}

\def\1{\bm{1}}

\DeclareMathAlphabet{\mathsfit}{\encodingdefault}{\sfdefault}{m}{sl}
\SetMathAlphabet{\mathsfit}{bold}{\encodingdefault}{\sfdefault}{bx}{n}

\usepackage{hyperref}
\usepackage{url}
\usepackage{booktabs}
\usepackage{array}
\usepackage{amsmath}
\usepackage{amssymb}
\usepackage{graphicx}
\usepackage{multirow}
\usepackage{xcolor}
\usepackage{makecell}
\usepackage{mdframed}
\usepackage{enumitem}
\usepackage{pifont}
\usepackage{wrapfig}
\usepackage{etoolbox}
\usepackage{fancyhdr}

\usepackage[most]{tcolorbox}

\newtcolorbox{promptbox}{
    breakable,
    colback=gray!5,
    colframe=gray!30,
    boxrule=0.5pt,
    arc=2pt,
    left=6pt,
    right=6pt,
    top=6pt,
    bottom=6pt,
    fontupper=\small\ttfamily,
    before skip=6pt,
    after skip=6pt
}

\makeatletter
\patchcmd{\WF@putfigmaybe}{\setbox\WF@box\hbox{\lower\intextsep\box\WF@box}}%
  {\global\advance\WF@size-\intextsep}{}{\PackageWarning{main}{wrapfig alignment patch failed}}
\let\WF@rapt@hdr\WF@rapt
\def\WF@rapt[#1]#2{\WF@rapt@hdr[#1]{#2}%
  \def\@tempa{table}\ifx\@captype\@tempa
    \setlength{\abovecaptionskip}{0pt}\setlength{\belowcaptionskip}{4pt}\fi}
\makeatother

\newcommand{\sqlts}{\texttt{SQL2S}}
\newcommand{\stsql}{\texttt{S2SQL}}

\newcommand{\dmc}{\textsc{sql-orig-syn}}
\newcommand{\dvfive}{\textsc{sql-orig-prim}}
\newcommand{\dstsql}{\textsc{sql-orig-s2sql-syn}}
\newcommand{\model}{\textsc{HybridResearcher-4B}}
\newcommand{\framework}{\textsc{HybridResearcher}}

\title{HybridResearcher: Bridging the Structured-Unstructured Boundary for Deep Research Agents}

\iclrfinalcopy 
\author{Soyoung Yoon\textsuperscript{\textmd 1}\thanks{~Equal contribution. Author order was determined randomly.}, 
\quad Peiran Xu\textsuperscript{\textmd 2}\footnotemark[1], 
\quad Han Wang\textsuperscript{\textmd 3}\footnotemark[1], 
\quad Ruofan Wu\textsuperscript{\textmd 4}\footnotemark[1], 
\quad Xiaodong Yu\textsuperscript{\textmd{5}$\dagger$}\footnotemark[1], \\
\textbf{Feng Yan}\textsuperscript{4}, 
~\textbf{Mohit Bansal}\textsuperscript{3}, 
~\textbf{Debiao Li}\textsuperscript{2},  
~\textbf{Seung-Won Hwang}\textsuperscript{1}, 
~\textbf{Yuxiong He}\textsuperscript{5}, 
~\textbf{Zhewei Yao}\textsuperscript{5$\dagger$} \\
{}\textsuperscript{1}Seoul National University \quad
{}\textsuperscript{2}University of California, Los Angeles \quad \\
{}\textsuperscript{3}University of North Carolina at Chapel Hill \quad
{}\textsuperscript{4}University of Houston \quad \\
{}\textsuperscript{5}Snowflake AI Research}
\begin{document}
\maketitle
\fancyhead{}
\renewcommand{\headrulewidth}{0pt}
\begingroup
\renewcommand{\thefootnote}{\fnsymbol{footnote}}
\footnotetext[2]{~Project Leader.}
\endgroup

\begin{abstract}
While LLM agents are increasingly capable of gathering and synthesizing evidence from a single homogeneous source, such as structured tables or unstructured data, real-world analytical questions frequently span heterogeneous sources.
For example, answering one enterprise question may require both a result computed over an internal database and evidence found on the web.
Recent evaluations on benchmarks like HybridDeepResearch show that even frontier agents struggle when evidence must cross this boundary. While these benchmarks have exposed the limitation, methodologies to explicitly teach cross-source reasoning remain absent.
To address this, we introduce \model{}, a 4B agent trained to bridge evidence across the structured and unstructured boundary. 
To build it, we synthesize cross-source bridge trajectories by starting with validated single-source tasks—such as existing text-to-SQL or web search queries—and fabricating the missing source around a bridge entity drawn from the seed's own output. We retain a sample only when strict construction gates confirm that neither side alone suffices, ensuring the data requires true cross-source interdependence. Finally, we train on this curated dataset using supervised fine-tuning followed by reinforcement learning.
Evaluated on the existing HybridDeepResearch benchmark, \model{} improves over our base model, Qwen3.5-4B, by 30.0 points pass@8 and surpasses the far larger Qwen3.5-397B-A17B by 8.3 points, while also improving over its base model on established, held-out single-source benchmarks (BrowseComp and BrowseComp-Plus), demonstrating that this complex cross-source reasoning can be taught directly to small models without triggering catastrophic forgetting.

\end{abstract}

\section{Introduction}
\label{sec:intro}

LLM agents extend language models beyond closed-form reasoning by letting them interact with external data sources through tools~\citep{yao2023react,schick2023toolformer}. 
A capable agent can write and execute SQL against a database, navigate the open web to locate and synthesize evidence, or query domain-specific APIs to retrieve information that parametric knowledge alone cannot supply. 
Training agents for these capabilities has advanced rapidly: recent recipes that combine synthetic data construction, supervised fine-tuning, and reinforcement learning have produced compact models that rival or surpass frontier models, in domains such as text-to-SQL~\citep{omnisql2025,yao-etal-2026-arctic}, and open-web deep research~\citep{jin2025searchr1,zheng2025deepresearcher,quest2026}. 
However, these recipes typically operate within a single homogeneous source, such as a database or the web, rarely bridging both in the same reasoning chain. Real-world analytical tasks rarely observe this boundary: domains routinely require synthesizing evidence across structured and unstructured sources to answer a single question—such as crossing internal transaction databases with public SEC filings for financial auditing, or merging inventory SQL with global news for supply-chain risk assessment~\citep{abaskohi2026drbench,fdabench2026,hdrbench2026}.
Recent benchmarks such as HybridDeepResearch~\citep{hdrbench2026} make this gap measurable, revealing that even frontier agents such as GPT-5~\citep{gpt5} and Claude-Sonnet-4.6~\citep{claude-sonnet-4-6} struggle to synthesize answers when evidence must cross this boundary. Yet while the evaluation gap is now visible, the training gap remains: to our knowledge, no prior work provides a training framework, training data, or a trained model for agents that must bridge structured and unstructured sources within a single reasoning chain.

To address this training gap, we ask: why does this capability gap persist despite rapid progress on each source in isolation? We identify two core obstacles. 
First, \textbf{component capabilities do not automatically compose into hybrid reasoning}. Fine-tuning a model on single-source corpora (whether SQL-only, search-only, or their union) without bridged trajectories fails to yield cross-source competence. While the union model almost fully recovers parallel tasks that exercise both skills independently, performance on genuinely directional tasks, where the output of one source conditions the other, remains severely impaired.
This demonstrates that hybrid data must explicitly teach the bridge itself: the interdependent execution between sources rather than the isolated skills. 
Second, naive hybrid synthesis is \textbf{vulnerable to shortcut learning}. 
Among teacher trajectories accepted by standard correctness filters, 79.5\% never execute a substantive SQL query. Instead, the model answers directly from parametric memory, bypassing the cross-source hop entirely. 
A model trained on such flawed trajectories merely learns to fake the bridge, a failure mode that scaling uncurated data only exacerbates. Overcoming both obstacles by explicitly teaching the cross-source bridge while enforcing strict anti-shortcut guarantees forms the foundation of our methodology.

\begin{figure}[t]
  \centering
  \includegraphics[width=0.9\linewidth]{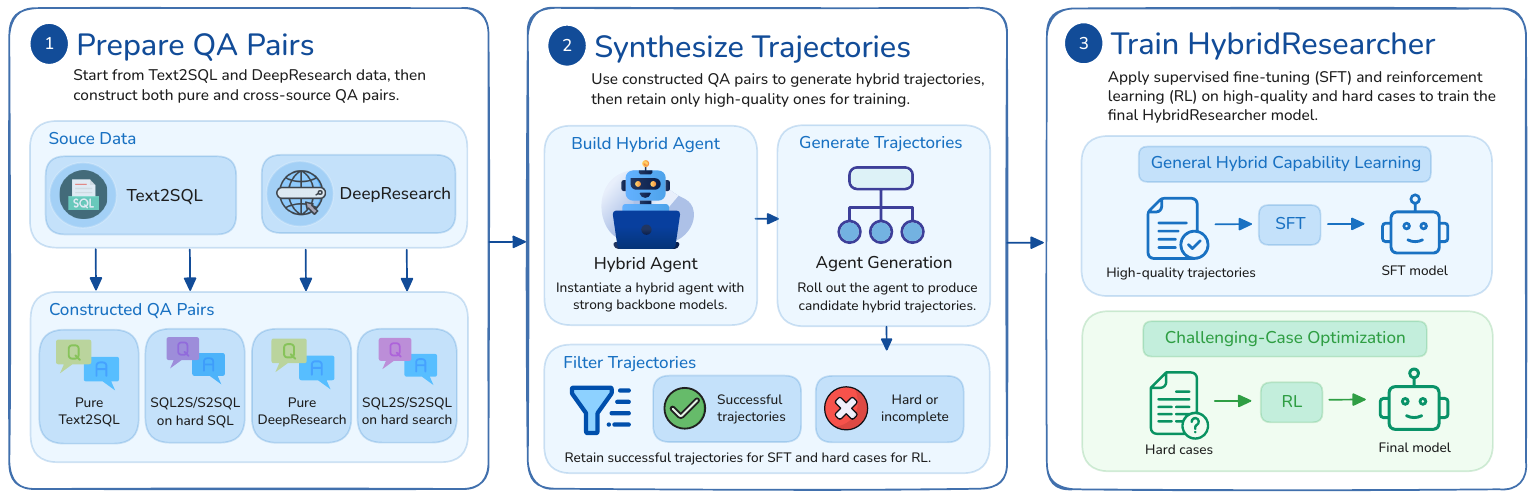}
  \caption{We start from validated single-source text-to-SQL and deep-research seeds, construct both single-source and cross-source (SQL2S/S2SQL) QA pairs, and use a hybrid agent to generate candidate trajectories. Successful trajectories are retained for supervised fine-tuning, while hard or incomplete cases are used for reinforcement learning, yielding \textsc{HybridResearcher}.}
  \label{fig:method_overview}
\end{figure}

We introduce \framework{}, a training framework that teaches hybrid deep research to small models, by extending validated single-source tasks across the structured/unstructured boundary. 
The core idea is that hybrid training data need not be fabricated from scratch. 
Each instance starts from a seed whose difficulty is already established—an existing text-to-SQL task or a web-search task. We synthesize the missing source around a bridge entity drawn from the seed's own output, so that the seed's difficulty is inherited rather than re-established. 
To directly address the shortcut problem, we retain an instance only when strict construction gates deterministically verify that neither source alone suffices, ensuring the bridge entity forms the unique, necessary path between them (examples in Appendix \ref{app:sql:examples} and \ref{app:searchorig:examples}).
While we instantiate this framework for the SQL–web pairing, the structural recipe---requiring only a validated single-source task to seed from and an entity that can bridge across boundaries---provides a clear blueprint for extending agentic reasoning to other heterogeneous combinations in the future.
The curated data serves both supervised fine-tuning and reinforcement learning, trained in a reproducible offline environment built around a frozen corpus and local retriever. To optimize this complex policy, our RL recipe introduces a majority-voting judge ensemble to mitigate reward noise, alongside strict hard-gating mechanisms that terminate and penalize degenerate, brute-force tool usage. Figure~\ref{fig:method_overview} summarizes this pipeline.

Evaluated on the HybridDeepResearch benchmark~\citep{hdrbench2026}, \model{} improves over its Qwen3.5-4B~\citep{qwen3.5} base model by 30.0 points pass@8 and surpasses the far larger Qwen3.5-397B-A17B-FP8 by 8.3 points. Crucially, this improvement does not come at the cost of single-source ability: \model{} also outperforms its base model on the held-out SQL-only benchmark Spider 2.0-lite~\citep{lei2025spider2} and on the web-search benchmarks BrowseComp~\citep{wei2025browsecomp} and BrowseComp-Plus~\citep{chen2025browsecompplusfairtransparentevaluation}. By explicitly teaching the cross-source bridge, our framework successfully imparts advanced hybrid reasoning capabilities to a compact model without triggering catastrophic forgetting~\citep{luo2025catastrophic}.

\section{Related Work}
\label{sec:related}

\textbf{Single-source Agents.}
Tool-using LLM agents have made rapid progress on tasks that draw evidence from one type of source.
For databases, benchmarks have moved from single-turn semantic parsing~\citep{yu2018spider,li2023bird} to realistic workflows over large enterprise schemas~\citep{lei2025spider2,livesqlbench2025}, and agents now explore the database and revise their queries based on execution results~\citep{deng2025reforce}. 
Small models trained on synthetic data~\citep{omnisql2025} or with reinforcement learning from execution rewards~\citep{ma2025sqlr1,yao-etal-2026-arctic,liu2025skyrlsql} are now competitive with much larger models on these benchmarks. 
For the web, search agents interleave reasoning with retrieval~\citep{yao2023react,jin2025searchr1,song2025r1searcher,zheng2025deepresearcher}, and recent systems can answer long-horizon questions whose answers are hard to find, as measured by BrowseComp~\citep{wei2025browsecomp}. 
These systems are trained largely on synthetic tasks designed to be hard, for example by obscuring key entities~\citep{li2025websailor}, formalizing the structure of information-seeking tasks~\citep{webshaper2025}, sampling paths from knowledge graphs~\citep{deepdive2025}, or generating fully synthetic research tasks~\citep{quest2026}, and then refined with reinforcement learning~\citep{gao2025asearcher}.
However, in all of these works, the agents work with a single source. 
In contrast, we train an agent that crosses the boundary between structured and unstructured sources, carrying evidence found in one source into its search of the other.

\textbf{Reasoning across Heterogeneous Sources.}
Hybrid question answering has long studied questions that require both tables and text~\citep{chen2020hybridqa,chen2021ottqa,chen2021finqa,zhu2021tatqa}, but the evidence is either given or retrieved in one step, and the model neither executes queries nor decides which source to consult next.
Other work adds language-model reasoning to database queries~\citep{liu-etal-2024-suql}.
Recent benchmarks evaluate agents that combine evidence from heterogeneous sources, including databases and documents~\citep{fdabench2026,ma2026dab,li2026dataspace}, enterprise files and the web~\citep{abaskohi2026drbench}, web pages and knowledge-graph APIs~\citep{yang2024crag}, databases and web search~\citep{hdrbench2026}, and multimodal evidence on the web~\citep{wang2026merrin}.
These benchmarks measure how well agents combine sources, but few works train agents to do so.
DSL-R1~\citep{hu2026dslr1} merges structured filters and vector search into a single query language over a closed collection, so the agent never has to carry a result from one kind of source to another. \citet{wubridging} synthesize simple two-hop questions that pair a SQL query with a single Wikipedia passage lookup, and train agents with hint-guided reinforcement learning. We instead present a full training recipe---from data construction to supervised fine-tuning and reinforcement learning---for significantly more challenging, multi-hop questions that combine hard SQL execution with long-horizon web search.

\section{Methodology: Synthesizing Cross-Source Trajectories}
\label{sec:methodology}

To train agents for hybrid reasoning, we require trajectories that exhibit true cross-source interdependence. 
Training on existing single-source data does not produce this capability, and synthesizing hybrid data naively invites shortcuts (\S\ref{sec:method:obstacles}). 
We therefore build hybrid trajectories from validated single-source seeds, starting from either text-to-SQL tasks (\S\ref{sec:data:sqlorig}) or search tasks (\S\ref{sec:data:searchorig}).

\subsection{Two Obstacles to Training Hybrid Agents}
\label{sec:method:obstacles}

\textbf{Single-Source Data Does Not Teach Hybrid Capability.}
Text-to-SQL~\citep{yu2018spider,li2023bird} and web search~\citep{du2026openseeker,chu2026redsearcher} training data are available in large quantities, whereas hybrid trajectories have to be synthesized. 
We therefore first test whether single-source data alone is enough. 
We fine-tune Qwen3.5-4B on successful trajectories from text-to-SQL tasks only, web-search tasks only, or their union, and evaluate on the hard subset of HybridDeepResearch \citep{hdrbench2026}, whose 120 questions each require both SQL and web search. None of the three mixtures helps. 
SQL-only training reaches 31.67 pass@8, within one question of the base model's 30.83, while union and search-only training fall to 29.17 and 15.83. 
Single-source trajectories never use the output of one tool as the input to the other, so training on them does not teach this step.

\textbf{Naive Hybrid Synthesis Invites Shortcuts.} 
Synthesizing hybrid questions and keeping the teacher trajectories that reach the correct answer is also not enough, because a correct answer does not show that both sources were used. 
When the facts on the search side are well known, the teacher can recall them instead of searching, and when the question reveals the bridge entity, the teacher can skip the hop that was meant to produce it. 
Models tend to trust such parametric knowledge even when it conflicts with retrieved evidence \citep{xie2024knowledgeconflicts,wang-etal-2025-adacad}. 
This risk is concrete in our setting. 
Bridge entities drawn from real databases are often famous, and the model's prior can override what it retrieves. 
Such trajectories lead the model to answer without crossing the boundary.

\begin{figure}[t]
\begin{center}
\includegraphics[width=\textwidth]{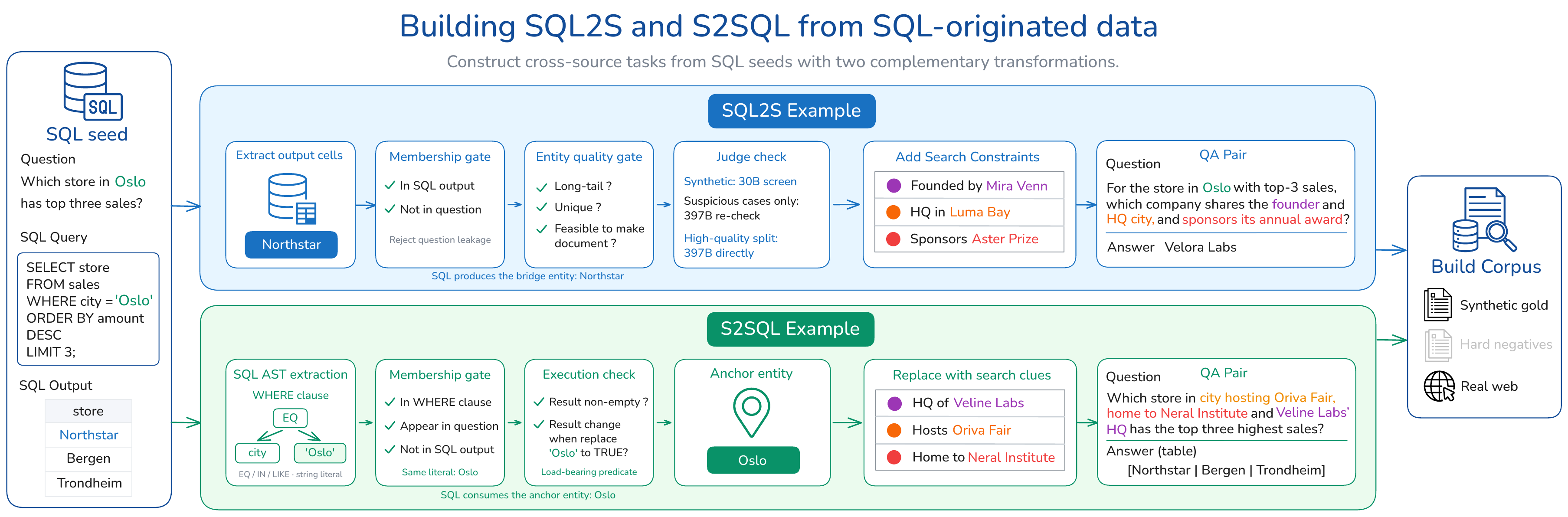}
\end{center}
\caption{Overview of sql-originated data construction.\textbf{Top (\sqlts{}):} an output cell (\emph{Northstar}) that passes
the membership gate, entity-quality gate, and judge check becomes the bridge; coined facts about it (HQ city ...) become search constraints that only one hub, \emph{Velora Labs}, satisfies. \textbf{Bottom (\stsql{}):} a \texttt{WHERE} literal from the SQL AST (\emph{Oslo}) that passes the membership gate and execution check becomes the anchor, and the answer is the unchanged SQL output. 
Gold and hard-negative documents for every clue are inserted with real web text by the retriever used at evaluation. (Details: \S\ref{sec:data:sqlorig} and Appendix~\ref{app:sql_data})}
\label{fig:sql_originated}
\end{figure}

\subsection{SQL-Originated Trajectories}
\label{sec:data:sqlorig}
We construct hybrid tasks from complete text-to-SQL seeds by deriving a multi-constraint web-retrieval subtask from a seed entity (Figure \ref{fig:sql_originated}, Appendix \ref{app:sql_data}).

\textbf{Building SQL2S.} \textbf{(i) Bridge Selection:} We select a cell from the executed output table to serve as the bridge entity (e.g., \emph{Northstar} in Figure~\ref{fig:sql_originated}). A membership gate ensures this cell appears in the output but not in the original question, deterministically forcing the agent to execute the SQL query to discover it. \textbf{(ii) Hop One (Coined Facts):} To prevent the model from bypassing the tool boundary via parametric memory, we invent novel facts (e.g., business rules) about the bridge entity using pronounceable pseudo-words. Each fact is paired with a fabricated gold document and typed hard negatives that perturb exactly one constraint. \textbf{(iii) Hop Two (The Intersection Hub):} The solver must resolve the bridge by SQL, resolve each coined fact by search, and find the single hub that all facts point to (e.g., \emph{Velora Labs}). The gold hub satisfies all $c$ constraints, while each near-miss distractor satisfies exactly $c-1$, forcing rigorous multi-document cross-referencing. \textbf{(iv) Verification:} The item must pass deterministic gates (e.g., ensuring the constraint intersection is a singleton) and model-based checks, including a mask test proving the answer cannot be guessed from negative documents.

\textbf{Building S2SQL.} \textbf{(i) Anchor Selection:} We select a string literal from an equality, inclusion, or pattern-matching predicate in the gold SQL query to serve as the anchor (Appendix~\ref{app:sql:examples}, \emph{Oslo} in Figure~\ref{fig:sql_originated}). An execution test requires that removing the predicate changes the result, and membership constraints require the literal to occur in the original question but not in the output. Together, these checks verify the anchor is load-bearing: the agent must resolve it via web search. \textbf{(ii) Relation Fan-out \& Hub Construction:} We coin multiple distinctive entities related to the anchor. The anchor is hidden in the final question and replaced with intersecting search clues. The agent is strictly forced to resolve these clues on the web to identify the anchor before it can formulate the correct SQL query.

\begin{figure}[t]
  \centering
  \includegraphics[width=\linewidth]{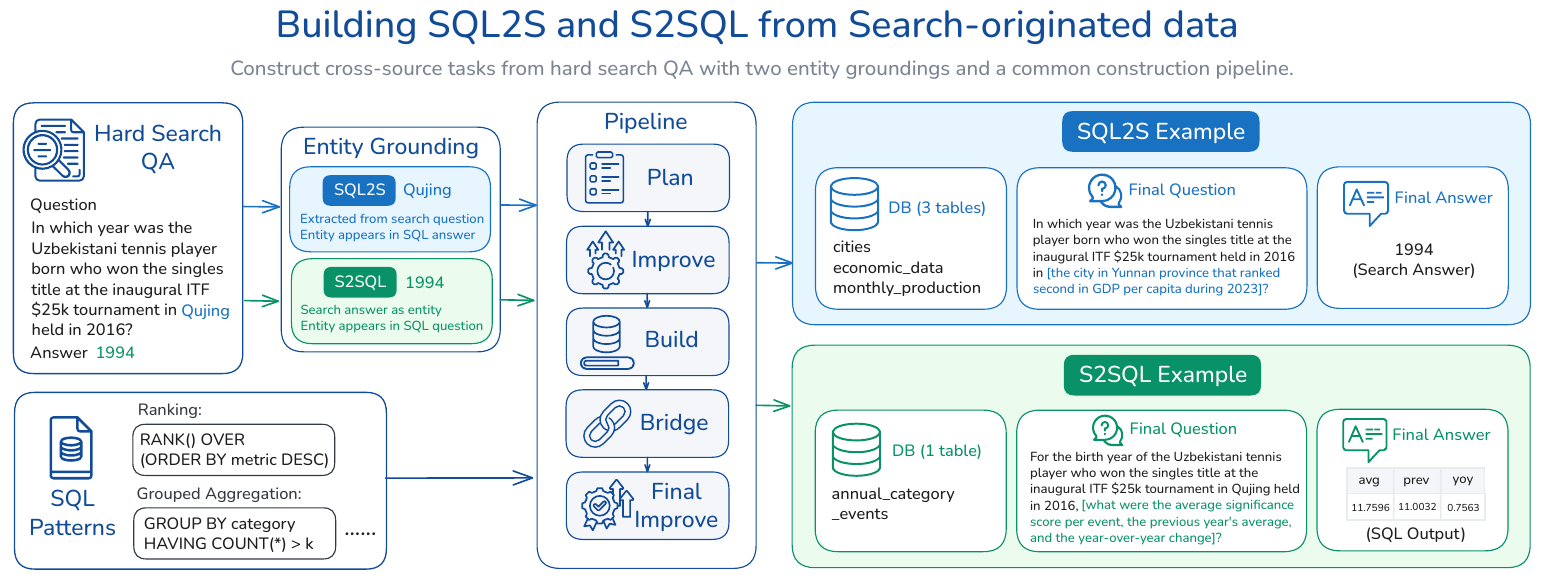}
  \caption{The search-originated pipeline. One hard-search item yields SQL2S or S2SQL according to whether the bridge is taken from the question or the answer. Both directions share the same stages: plan, improve, build, bridge, and a final improve, and draw SQL difficulty from a catalog of executable patterns.}
  \label{fig:search_orig}
\end{figure}

\subsection{Search-Originated Trajectories}
\label{sec:data:searchorig}
We construct hybrid tasks from hard-search seeds by adding a synthetic text-to-SQL component around a seed entity (Figure \ref{fig:search_orig}, Appendix~\ref{app:searchorig}). The original seed supplies the web-retrieval requirement; the constructed database supplies the independently nontrivial SQL requirement. A single search item can support both directions (SQL2S, S2SQL) depending on which text span is selected as the bridge.

\textbf{SQL2S:} The bridge is an entity mentioned in the original search question (e.g., \emph{Qujing} in Figure \ref{fig:search_orig}). We formulate a SQL task where this entity is the answer. The agent is forced to query the database to discover the entity before it can complete the web search.

\textbf{S2SQL:} The bridge is the search answer itself (e.g., \emph{1994}). This value becomes a required parameter that the SQL query consumes. The agent must successfully resolve the web search to find this parameter before executing the database query.

\textbf{Construction Pipeline.} Both directions run through a multi-stage generation pipeline to ensure the fabricated database requires genuine structured reasoning, preventing it from degrading into a trivial lookup table:
\textbf{(i)} \textbf{Plan:} We select advanced SQL techniques from a catalog of executable patterns (e.g., grouped aggregation, ranking) and design a schema containing only raw observational data (e.g., individual event records rather than precomputed yearly stats).
\textbf{(ii)} \textbf{Improve:} The schema is refined to strictly eliminate any columns that might pre-aggregate the answer, logically guaranteeing the gold query cannot be simplified by shortcut logic.
\textbf{(iii)} \textbf{Build:} The planned schema is materialized into a populated SQLite database. To prevent trivial edge-case filtering, we dynamically generate same-kind peer entities and bury the target entity deeply within the data distribution.
\textbf{(iv)} \textbf{Bridge \& Final Improve:} The original search clues and the newly fabricated SQL constraints are merged into a single cohesive prompt. The bridge entity is completely masked, and scope hints are added only to resolve genuine ambiguities.

\textbf{Load-Bearing Verification.} To guarantee interdependence, both hops must pass strict execution checks. For SQL2S, the bridge must never appear in the merged question text. For S2SQL, the bridge must not leak into the final SQL result table. Crucially, the fabricated database must present a structurally fresh query rather than merely mirroring the search clues; otherwise, the SQL tool acts as a trivial lookup cache for a web puzzle the agent has already solved.

\section{Training Recipe}

\subsection{Supervised Fine-tuning}
\textbf{Trajectory generation.} Our SFT mixture combines directly used training data (Appendix~\ref{app:direct_data}) with judge-correct trajectories generated by running the agent harness (Appendix~\ref{app:harness}) against the local corpus (Appendix~\ref{sec:app:local_corpus}) and databases. We use either Qwen3.5-397B-A17B-FP8~\citep{qwen3.5} or DeepSeek-V4-Flash~\citep{deepseekai2026deepseekv4} for generating trajectories. We set temperature $0.6$, top-$p$ $0.95$, top-$k$ $20$, thinking on, $3600$\,s timeout, with $32k$ max tokens set to SQL-originated tasks and $128k$ for search-originated tasks. Max turns range from $100$ to $300$ depending on the task.

\subsection{Reinforcement Learning}
\label{sec:main_rl}

\paragraph{Data Curation.} We collect 21K synthesized hybrid tasks that remained unsolved during the SFT stage as candidates for RL. Because RL is highly sensitive to noisy data~\citep{ye2026codemidasscalingagenticcoding, tang2025highdataefficiencyreinforcement}, we aggressively filter this pool down to 0.7K high-signal QA pairs using a strict three-stage pipeline. 

\textbf{(i) Difficulty filtering:} we execute preliminary rollouts and discard tasks that are either consistently solved (too easy) or never solved (too hard or unsolvable), ensuring the dataset targets the exact frontier of the model's capabilities. 

\textbf{(ii) Quality filtering:} a data-quality agent removes QA pairs with ambiguous answers, duplicate search chains, or inconsistent cross-source evidence. We actively prioritize questions requiring broad multi-hop retrieval or complex SQL operations. 

\textbf{(iii) Distribution balancing:} to prevent policy collapse on a single source or task family, we strictly balance the final 0.7K dataset 1:1 between SQL-originated and search-originated tasks, and 1:1 between the \stsql{} and \sqlts{} directions.

\paragraph{RL Recipe.} Our RL training is based on CISPO \citep{minimaxm1}, with several techniques introduced to improve training efficiency and stability. Further details are provided in Appendix~\ref{app:rl-recipe}.

\textbf{Hard-gating:} 
As invalid tool interactions and malformed tool calls will introduce noisy or substantial variance training signals into policy optimization, we applied strict hard-gating rules to increase training and tool-use efficiency: A rollout is terminated early and assigned zero reward when it exhausts the turn budget without making a tool call, contains malformed tool arguments, invokes nonexistent tools, fails to produce a valid final answer, or repeatedly issues duplicated tool queries.

\textbf{Judge majority-voting:} 
RL is highly sensitive to noisy reward signals. Relying on a single LLM judge frequently produces unreliable evaluations, particularly when correctness requires nuanced semantic matching (e.g., distinguishing \emph{Charles III} from \emph{Charles Emmanuel III}) or recognizing semantically equivalent SQL solutions that differ in surface-level table form. To stabilize optimization, both \stsql{} and \sqlts{} use an LLM-as-a-judge with a five-vote majority to determine the binary pass/fail reward. An empirical analysis of this voting mechanism is presented in \S\ref{sec:judge-vote-analysis}.

\section{Evaluation}
\label{sec:eval}

\textbf{Benchmarks and metrics.}
We evaluate hybrid reasoning on the 120-question hard subset of HybridDeepResearch~\citep{hdrbench2026}, evenly split across Search-to-SQL (S2SQL), SQL-to-Search (SQL2S), and Parallel tasks. To measure whether hybrid training preserves the underlying component capabilities, we measure single-source retention on the 1,266 BrowseComp~\citep{wei2025browsecomp} full set and 830 queries of BrowseComp-Plus~\citep{chen2025browsecompplusfairtransparentevaluation} full set for web search, and the 135 public subset of Spider~2.0-lite~\citep{lei2025spider2} for text-to-SQL. BrowseComp is evaluated against the live web, whereas BrowseComp-Plus uses its fixed corpus of 100,195 documents, providing complementary evaluation under dynamic and static retrieval environments. None of these benchmarks appear in our training data. Additional benchmark and evaluation details are provided in Appendix~\ref{app:eval}. 

\textbf{Protocol \& Environment.} For hybrid and search-only tasks, we evaluate the agent's final answers using a Qwen3.8-27B~\citep{qwen38} LLM judge with a 5-vote majority to ensure reward reliability (see \S\ref{sec:judge-vote-analysis}). Spider 2.0-lite relies on its official execution-accuracy metric. Models interact with the SQL and retrieval environments through our custom agent harness (Appendix~\ref{app:harness}). A comparison with other agent harnesses is in Appendix~\ref{app:harness:comparison}. 
Unless a benchmark specifies its own retrieval setting, the web tools are served from the same local corpus used in training (Appendix~\ref{sec:app:local_corpus}).
BrowseComp-Plus~\citep{chen2025browsecompplusfairtransparentevaluation} restricts retrieval to a fixed 100K-document corpus, while standard BrowseComp~\citep{wei2025browsecomp} operates against the live web. Unless noted otherwise, we report avg@8 and pass@8.

\subsection{Results}
\label{sec:eval:results}

\paragraph{Hybrid training progressively unlocks cross-source reasoning.} As shown in Table~\ref{tab:training_progression}, the Qwen3.5-4B base model fundamentally struggles with cross-source tasks, achieving only 30.83\% overall pass@8. Our hybrid SFT framework successfully bridges this gap, nearly doubling the performance to 59.17\%. Applying RL yields further gains, particularly in the avg@8 metric (increasing from 23.75\% to 27.19\% overall). This improvement highlights that RL makes the model more confident and consistent in arriving at the correct answer, primarily by strengthening the strictly interdependent directional tasks (e.g., S2SQL avg@8 increases from 26.88\% to 31.56\%). Remarkably, despite its small size, our final 4B agent approaches the performance of substantially larger frontier models like Claude Sonnet 4.6 (65.83\% pass@8), GLM-5.2-FP4~\citep{glm5} (63.33\% pass@8) and decisively surpasses its own family's flagship, Qwen3.5-397B-A17B-FP8 (52.50\% pass@8 and 26.77\% avg@8), on both metrics.

\begin{table}[t]
\centering
\small
\caption{Hybrid training progression and comparison with larger models (Pass@8 and Avg@8, \%).}
\label{tab:training_progression}
\resizebox{0.95\textwidth}{!}{
\begin{tabular}{lcccc|cccc}
\toprule
& \multicolumn{4}{c|}{\textbf{Pass@8}} & \multicolumn{4}{c}{\textbf{Avg@8}} \\
\textbf{Model} & \textbf{Overall} & \textbf{S2SQL} & \textbf{SQL2S} & \textbf{Parallel}
& \textbf{Overall} & \textbf{S2SQL} & \textbf{SQL2S} & \textbf{Parallel} \\
\midrule
Qwen3.5-397B-A17B-FP8
& 52.50 & 52.50 & 32.50 & 72.50
& 26.77 & 32.50 & 13.75 & 34.06 \\
GLM-5.2-FP4
& 63.33 & 65.00 & 47.50 & 77.50
& \textbf{38.54} & 32.19 & 27.50 & 55.94 \\
Claude Sonnet 4.6
& \textbf{65.83} & 57.50 & 52.50 & 87.50
& 37.29 & 30.31 & 36.25 & 45.31 \\
\midrule
Qwen3.5-4B
& 30.83 & 22.50 & 20.00 & 50.00
& 7.71 & 5.00 & 3.44 & 14.69 \\
+ Hybrid SFT
& 59.17 & 65.00 & \textbf{35.00} & 77.50
& 23.75 & 26.88 & 8.75 & 35.83 \\
+ Hybrid RL
& \textbf{60.83} & \textbf{70.00} & 30.00 & \textbf{82.50}
& \textbf{27.19} & \textbf{31.56} & \textbf{12.81} & \textbf{37.19} \\
\bottomrule
\end{tabular}
}
\end{table}

\begin{table}[t]
\centering
\small
\caption{Single-source performance before and after hybrid training (\%): search-only on BrowseComp and
BrowseComp-Plus, SQL-only on the held-out Spider~2.0-lite subset (execution accuracy).}
\label{tab:search_only}
\begin{tabular}{@{}lcccccc@{}}
\toprule
& \multicolumn{2}{c}{\textbf{BrowseComp}}
& \multicolumn{2}{c}{\textbf{BrowseComp-Plus}}
& \multicolumn{2}{c}{\textbf{Spider~2.0-lite}} \\
\cmidrule(lr){2-3}
\cmidrule(lr){4-5}
\cmidrule(l){6-7}
\textbf{Model}
& \textbf{Pass@8} & \textbf{Avg@8}
& \textbf{Pass@8} & \textbf{Avg@8}
& \textbf{Pass@8} & \textbf{Avg@8} \\
\midrule
Qwen3.5-4B
& 54.03 & 29.57
& 60.84 & 36.49
& 59.26 & 30.65 \\
+ Hybrid SFT
& 61.37 & 36.79
& 69.76 & 46.36
& \textbf{65.19} & \textbf{41.39} \\
+ Hybrid RL
& \textbf{62.24} & \textbf{38.90}
& \textbf{73.13} & \textbf{50.56}
& 62.96 & 40.83 \\
\bottomrule
\end{tabular}
\end{table}

\paragraph{Hybrid training preserves and improves single-source capabilities.} 
Fine-tuning a compact model on complex new behaviors frequently risks catastrophic forgetting of component skills. However, Table~\ref{tab:search_only} demonstrates that \model{} entirely avoids this tradeoff. Instead of cannibalizing existing knowledge, hybrid SFT actually creates a synergistic effect, substantially improving the base model's standalone capabilities. On the search side, BrowseComp pass@8 increases from 54.03\% to 61.37\%, and BrowseComp-Plus from 60.84\% to 69.76\%. On the SQL side, Spider 2.0-lite execution accuracy jumps from 59.26\% to 65.19\%. Applying hybrid RL largely preserves these single-source improvements (e.g., further boosting BrowseComp-Plus to 73.13\% pass@8). This confirms that explicitly teaching the cross-source bridge reinforces, rather than replaces, the agent's core reasoning foundation.

\section{Analysis}
\label{sec:analysis}

\subsection{Behavioral Shifts: How SFT and RL Transform Tool Use}
\label{sec:analysis:sft_rl}

\begin{table}[t]
\begin{minipage}[t]{0.52\textwidth}
\caption{Search behavior on the evaluation set ($960$ rollouts per model; \stsql{} and \sqlts{} rows use their
$320$). \sqlts{} accuracy is conditioned on the gold answer appearing in a search result ($n = 43/59/79$
rollouts) or anywhere on the web side ($n = 85/130/147$). $\Delta$: relative change SFT$\rightarrow$RL.}
\label{tab:search_overview}
\centering\footnotesize
\setlength{\tabcolsep}{3pt}
\resizebox{\textwidth}{!}{
\begin{tabular}{@{}lrrrr@{}}
\toprule
 & \textbf{Base} & \textbf{SFT} & \textbf{RL} & $\Delta$ \\
\midrule
\multicolumn{5}{@{}l}{\emph{Search and reading}} \\
\quad Web searches / rollout & 38.6 & 22.1 & 18.7 & $-$15\% \\
\quad Page visits / rollout & 7.0 & 9.9 & 14.3 & $+$45\% \\
\quad Page visits / search & 0.29 & 0.50 & 0.88 & $+$76\% \\
\midrule
\multicolumn{5}{@{}l}{\emph{Evidence}} \\
\quad \stsql{}: bridge surfaced (\%) & 69.7 & 70.9 & 77.8 & $+$10\% \\
\quad \sqlts{}: acc.\ $\mid$ gold surfaced & 12.9 & 18.5 & 27.2 & $+$47\% \\
\bottomrule
\end{tabular}}
\end{minipage}\hfill
\begin{minipage}[t]{0.45\textwidth}
\caption{Search queries and efficiency of the SFT and RL models ($960$ rollouts each). Near repeats measure repeats of an earlier query (Jaccard $>=$ 0.8). $\Delta$ as in
Table~\ref{tab:search_overview}.}
\label{tab:search_queries}
\centering\footnotesize
\setlength{\tabcolsep}{3pt}
\resizebox{\textwidth}{!}{
\begin{tabular}{@{}lrrr@{}}
\toprule
 & \textbf{SFT} & \textbf{RL} & $\Delta$ \\
\midrule
\multicolumn{4}{@{}l}{\emph{Queries}} \\
\quad Words / query & 6.0 & 6.5 & $+$9\% \\
\quad Characters / query & 46.9 & 52.3 & $+$11\% \\
\quad With a quoted phrase (``  ") & 41.4 & 48.8 & $+$18\% \\
\quad With operator (site:, OR, -term) & 0.6 & 1.2 & $+$126\% \\
\quad Exact repeats / rollout & 0.63 & 0.23 & $-$63\% \\
\quad Near repeats / rollout & 4.09 & 2.77 & $-$32\% \\
\midrule
\multicolumn{4}{@{}l}{\emph{Efficiency}} \\
\quad Turns / rollout & 44.7 & 42.0  & $-$6\% \\
\quad Tool calls / turn & 1.28 & 1.39 & $+$9\% \\
\quad Output tokens / rollout (K) & 14.8 & 14.1 & $-$5\% \\
\bottomrule
\end{tabular}
}
\end{minipage}
\end{table}

To understand how our training recipe builds hybrid capability, we analyze the agent's search behavior on the evaluation set (Tables~\ref{tab:search_overview} and \ref{tab:search_queries}).

\textbf{SFT teaches structured exploration.} The base model exhibits a naive search policy, firing off many queries (38.6 per rollout) but rarely reading the results (0.29 page visits per search). SFT corrects this imbalance. The agent learns to systematically explore retrieved content, nearly halving the number of web searches while significantly increasing page visits. This transition from blind searching to active reading improves basic evidence extraction, raising \sqlts{} accuracy (when the gold answer is present) from 12.9\% to 18.5\%.

\textbf{RL optimizes for precision and evidence utilization.} While SFT teaches the basic mechanics, RL concentrates the policy on its successful behaviors. After RL, the agent's search behavior becomes highly deliberate: queries become longer, more specific (48.8\% use quotes, and operator use doubles), and far less redundant (exact repeats drop by 63\%). Consequently, the agent extracts evidence effectively; on \sqlts{}, the agent's ability to capitalize on surfaced gold answers jumps from 18.5\% to 27.2\%. Crucially, RL achieves these gains without inflating the trajectory cost, maintaining similar or less turn counts and token usage to the SFT checkpoint.

\textbf{Remaining Bottlenecks.} Despite these improvements, cross-source reasoning remains bottlenecked by retrieval limits and early stopping. On \stsql{}, the required bridge entity still fails to surface in 22.2\% of rollouts. Moreover, 82.0\% of RL's wrong answers (vs.\ 48.7\% of its correct ones) cites a time budget, the effort already spent, or a guess (Appendix~\ref{app:failure:uncertain}), suggesting they commit under self-acknowledged uncertainty. Making the decision to search more when uncertain and refraining from guessing could be a promising next direction.

\subsection{Data Quality Outweighs Scale in RL Optimization}
\label{sec:analysis:data_quality}

To demonstrate that RL benefits more from high-signal data than sheer volume, we compare our curated RL dataset against the full 21K candidate pool using the identical training recipe. Despite being a fraction of the size, the curated dataset achieves a 27.2\% overall avg@8, substantially outperforming the massive unfiltered pool (21.0\%). This performance gap is driven by training efficiency: in the curated dataset, 71\% of rollout groups provide sufficient reward contrast to contribute to policy updates, compared to just 38\% in the uncurated pool. This confirms that aggressive filtering prevents noisy or uninformative trajectories from diluting the RL optimization process, making data curation a much stronger lever than scaling data volume.

\subsection{Judge majority-voting helps improve reward reliability}
\label{sec:judge-vote-analysis}

To quantify the impact of the majority judging ensemble introduced in Section~\ref{sec:main_rl}, we compare it against a single-vote baseline. The five-vote mechanism improves the model's overall avg@8 from 25.2\% to 27.2\%. Telemetry from our training rollouts reveals exactly why this stabilization is necessary: a single judge's verdict disagrees with the final consensus in 8.2\% of cases. This instability is heavily concentrated in the \stsql{} direction (12.2\% disagreement, compared to just 1.2\% for \sqlts{}), confirming that evaluating executable logic combined with complex web entities is highly variance-prone. By smoothing out these individual judge errors (examples in Appendix~\ref{app:judge_examples}), majority voting successfully prevents noisy rewards from degrading the policy.

\subsection{Hard-Gating yields cleaner training signal and efficient tool-use}
\begin{figure}[t]
  \centering
  \includegraphics[width=\linewidth]{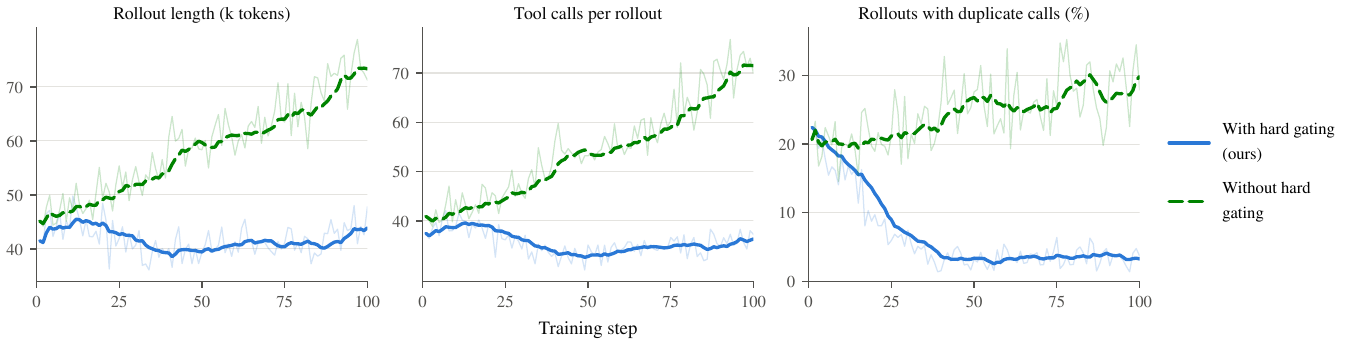}
  \caption{Training rollout length (left), tool calls per rollout (middle) and the share of rollouts with duplicated tool calls (right), with and without hard gating. Hard gating helps suppress redundant tool calls and stabilize rollout length.}
  \label{fig:hardgate}
\end{figure}

Hard-gating termination substantially improves both performance and rollout efficiency. Removing the hard-gating rules lowers avg@8 from 27.2\% to 25.6\%. We attribute this degradation to a positive feedback loop in tool use: without hard gating, redundant or invalid actions are insufficiently penalized, causing the policy to increasingly favor prolonged trajectories and redundant tool calls. As shown in Figure~\ref{fig:hardgate}, both trajectory token length and tool-call counts increase throughout training without hard gating, while remaining stable with hard gating. This trend also carries over to evaluation: hard gating reduces average tool calls from 76 to 55 per question and repeated-call frequency from 49\% to 11\%. The fraction of questions failing to produce a valid answer also drops from 10.7\% to 0.0\%. Overall, hard gating suppresses inefficient tool use and provides cleaner training signals, leading to more efficient and reliable tool use, consistent with the observations in \S\ref{sec:analysis:sft_rl}.

\section{Conclusion}
We studied how to train agents for questions that require evidence from both a structured database and unstructured web text, where a result obtained from one source must be carried into the other. 
Training on existing single-source data does not produce this capability, and synthesizing hybrid data naively invites shortcuts. 
We therefore build training data by extending validated text-to-SQL and deep-search tasks across the boundary, keep only instances that construction gates confirm require both sources, and train on them with supervised fine-tuning followed by reinforcement learning. 
\model{} improves over its base model by 30.00 points pass@8 on HybridDeepResearch and outperforms Qwen3.5-397B-A17B, while also improving on single-source SQL and search benchmarks. 
Our analysis shows that fine-tuning teaches the agent to read what it retrieves instead of issuing many searches, and that reinforcement learning benefits more from a small curated task set than from a larger unfiltered pool. 
More broadly, this work provides a complete training recipe for agents that reason across heterogeneous sources, and we hope it can be carried over to real-world settings that combine other kinds of sources, such as internal documents, code repositories, or domain-specific databases.

\newpage

\subsection*{AI use statement}

\textbf{Required disclosures.} Generative AI is part of our method, so we disclose several required uses.
(i)~\emph{Generating synthetic datasets.} The training data of \S\ref{sec:methodology} were synthesized with
LLMs, but every item starts from a real seed (a text-to-SQL query or a web-search question). SFT trajectories
were sampled from teacher models.
(ii)~\emph{Cleaning and filtering datasets.} LLM judges scored candidate entities, checked data quality, and ran
the model-based verification gates, alongside deterministic, non-LLM gates.
(iii)~\emph{Evaluation.} HybridDeepResearch, BrowseComp, and BrowseComp-Plus answers were graded by an LLM judge
(Qwen3.8-27B, majority of five votes); Spider~2.0-lite uses its deterministic execution-accuracy metric.

\textbf{Recommended disclosures.} We used generative AI tools to polish the paper for readability and to
brainstorm figure ideas; every figure in this version was drawn and edited by the authors.

\textbf{Not used.} We did not use generative AI to formulate mathematical claims, write proofs, or translate.
Surveys, interviews, and transcription are not applicable to this work.

\textbf{Verification.} We reviewed all AI-assisted work. Synthetic items pass deterministic gates, and the
authors inspected samples of every shipped subset. Every reported number was checked against the training logs,
evaluation outputs, or data files it came from.

We take responsibility for the final content of this work, including text, claims, and artifacts produced with
the aid of generative AI.

\bibliography{hdr}
\bibliographystyle{iclr2027_conference}

\clearpage
\appendix
\section{Overview}
\label{app:overview}

This appendix provides the data-construction details, environment, evaluation protocol, and extended analyses
behind the main text. Specifically:
\begin{itemize}[leftmargin=*,topsep=2pt,itemsep=1pt,parsep=0pt,partopsep=0pt]
\item Appendix~\ref{app:direct_data} lists the existing datasets we train on directly: HyDRA items and nine
public text-to-SQL datasets.
\item Appendix~\ref{app:sql_data} details the SQL-originated construction: the shipped subsets, the \sqlts{} and
\stsql{} builds, corpus ingestion, and worked examples.
\item Appendix~\ref{app:searchorig} details the search-originated construction: the seed pool and pipeline, the
shipped databases, and worked examples.
\item Appendix~\ref{app:sft_ablation} reports the SFT data ablation that adds each data family in turn.
\item Appendix~\ref{sec:app:local_corpus} describes the local retrieval corpus that serves the web tools during
trajectory generation, RL, and evaluation.
\item Appendix~\ref{app:harness} describes the hybrid agent harness: its tool interface, episode execution, and a
comparison with other agent harnesses.
\item Appendix~\ref{app:eval} gives the evaluation details for BrowseComp, BrowseComp-Plus, and Spider~2.0-lite,
the inference configuration of every model, and the LLM judge prompts.
\item Appendix~\ref{app:failure} analyzes the failures of the base, SFT, and RL models: the outcome of every
rollout and commits justified by time, effort, or a guess.
\item Appendix~\ref{app:rl} gives the RL training parameters and an analysis of judge voting.
\end{itemize}

\section{Direct Training Data}
\label{app:direct_data}

We first include existing datasets that are used directly for training rather than as construction seeds. Specifically, we use 10{,}184 HyDRA items~\citep{wubridging} and 25{,}747 deduplicated text-to-SQL examples drawn from nine public datasets: Spider 1.0~\citep{yu2018spider}, BIRD~\citep{li2023bird},
SynSQL~\citep{omnisql2025}, EHRSQL \citep{lee2022ehrsql},
LLMSQL~\citep{llmsql}, MMQA~\citep{wu2025mmqa}, LogicCat~\citep{liu2026logiccat},
SParC~\citep{yu-etal-2019-sparc}, and CoSQL~\citep{yu-etal-2019-cosql}. These examples provide direct supervision for hybrid and SQL capabilities, respectively. We use a random subset of SynSQL; for the other datasets, we keep only the questions for which at least one sampled trajectory is correct. Table~\ref{tab:sqlorig:base} summarizes the composition.

\begin{table}[h]
\caption{Base data: trained on directly. Qwen3.5 tokens at $128$k packing.}
\label{tab:sqlorig:base}
\begin{center}
\small
\resizebox{\textwidth}{!}{%
\begin{tabular}{@{}lrrrrrrrrrrr@{}}
\toprule
 & & \multicolumn{9}{c}{\emph{nine raw text-to-SQL sources} ($25{,}747$ rows, $115.7$M tokens)} & \\
\cmidrule(lr){3-11}
 & HyDRA & Spider~1.0 & BIRD & SynSQL & EHRSQL & LLMSQL3 & MMQA & LogicCat & SParC & CoSQL & \textbf{Total} \\
\midrule
Rows       & $10{,}184$ & $7{,}791$ & $6{,}231$ & $3{,}360$ & $3{,}107$ & $2{,}587$ & $1{,}942$ & $592$ & $130$ & $7$ & $\mathbf{35{,}931}$ \\
Tokens (M) & $196.1$ & $24.9$ & $24.3$ & $23.4$ & $23.7$ & $9.3$ & $6.1$ & $3.5$ & $0.5$ & $0.04$ & $\mathbf{311.7}$ \\
\bottomrule
\end{tabular}}
\end{center}
\end{table}

\section{SQL-Originated Construction Details}
\label{app:sql_data}

Specifications and counts behind \S\ref{sec:data:sqlorig}, organized by the construction steps of the main text.
Three SQL-originated subsets ship (Table~\ref{tab:sql:shipped}): \dmc{} and \dvfive{} hold \sqlts{} items
(Appendix~\ref{app:sql:sql2s}), and \dstsql{} holds \stsql{} items (Appendix~\ref{app:sql:anchor}).
Construction seeds are drawn from two source families summarized in Table~\ref{tab:sql:survey}: primary datasets with real databases and named entities, and a large synthetic SQLite source.

\begin{table}[h]
\caption{The three SQL-originated subsets that ship. Token counts are Qwen3.5 tokens at $128$k packing.}
\label{tab:sql:shipped}
\begin{center}
\small
\begin{tabular}{lrrl}
\toprule
\textbf{Subset} & \textbf{Rows} & \textbf{Tokens} & \textbf{Direction / split} \\
\midrule
\dmc{}     & $14{,}985$ & $598.3$M & \sqlts{}, synthetic \\
\dvfive{}  & $15{,}067$ & $272.2$M & \sqlts{}, primary \\
\dstsql{}  & $6{,}269$  & $272.8$M & \stsql{}, synthetic \\
\midrule
\textbf{total} & $\mathbf{36{,}321}$ & $\mathbf{1{,}143.3}$\textbf{M} & \\
\bottomrule
\end{tabular}
\end{center}
\end{table}

\begin{table}[h]
\caption{Seed-source for shipped SQL-originated data, and the reason for the selection.}
\label{tab:sql:survey}
\begin{center}
\small
\begin{tabular}{@{}>{\raggedright\arraybackslash}p{0.10\textwidth}>{\raggedright\arraybackslash}p{0.45\textwidth}>{\raggedright\arraybackslash}p{0.40\textwidth}@{}}
\toprule
\textbf{Family} & \textbf{Members} & \textbf{Verdict} \\
\midrule
\textsc{primary} & BIRD~\citep{li2023bird}, Spider~1.0~\citep{yu2018spider}, BULL/FinSQL~\citep{zhang2024finsql}, Archer~\citep{zheng-etal-2024-archer}, KaggleDBQA~\citep{lee-etal-2021-kaggledbqa}, Classics~\citep{finegan-dollak-etal-2018-improving} (Advising, ATIS~\citep{data-atis-original,data-atis-geography-scholar}, Geography~\citep{data-geography-original,data-atis-geography-scholar}, Scholar~\citep{data-atis-geography-scholar}, Restaurant~\citep{data-restaurants-logic,data-restaurants-original,data-restaurants}, IMDB~\citep{data-sql-imdb-yelp}) & real databases, real named entities; form the \textbf{primary} split \\
\midrule
\textsc{synthetic} & SynSQL-2.5M (OmniSQL; \citealp{omnisql2025}) & the only synthetic set that ships populated databases ($16{,}583$ SQLite DBs, $2.5$M English rows); forms the \textbf{synthetic} split \\
\bottomrule
\end{tabular}
\end{center}
\end{table}

\subsection{Building \sqlts{}: the bridge is an output cell}
\label{app:sql:sql2s}

\dmc{} (synthetic split) runs all four steps below; \dvfive{} (primary split) is then refabricated by an
anti-shortcut pass. Examples~1 and~2 in Appendix~\ref{app:sql:examples} walk through one item of each.

\begin{itemize}
    \item[\textbf{(i)}] \textbf{Bridge Selection:} The bridge is a cell of the executed gold output that the
    question does not name. An LLM judge scores each candidate on rarity, referent-collision risk, and whether a
    credible document can mention it; candidates that are ambiguous or cannot host a fabricated document are
    dropped.
    \item[\textbf{(ii)}] \textbf{Hop One (Coined Facts):} Each fact is built answer-first: the bridge is pinned
    from the gold SQL, a fact about it (a business rule, a data-dictionary definition, or a specification value)
    is invented, and only then is the question written.
    Coined names are pronounceable pseudo-words made globally unique by a reservation registry. Answers are
    \textsc{name} types (\texttt{coined\_company\_org}, \texttt{coined\_team}, \texttt{coined\_person},
    \texttt{coined\_product\_model}, \texttt{coined\_award}) or \textsc{value} (\texttt{number}).
    Questions may also define a coined composite metric (name, abbreviation, description, and a formula faithful
    to the gold SQL) in a domain-knowledge block: $77.4\%$ of \dmc{} and $14.0\%$ of \dvfive{} questions have
    one. Only \dmc{} also serves these definitions through the knowledge-lookup tools
    (Table~\ref{tab:harness_tools}), since the real databases behind \dvfive{} have no knowledge base. Each fact
    gets one gold document, which wraps the answer sentence in a realistic body in a register chosen by entity
    type (news or biography for people, press release or filing for organizations, spec page or review for
    products, internal wiki for private metrics); golds average $890$--$924$ characters, so most fit one
    retrieval chunk. Each fact also gets up to ten typed hard negatives, full-length perturbations of the gold's
    answer sentence that each flip exactly one constraint (Table~\ref{tab:sql:negatives}).
    \item[\textbf{(iii)}] \textbf{Hop Two (The Intersection Hub):} Given $n$ coined facts about the bridge, hubs
    are minted over fact subsets: the gold hub satisfies all $c$ constraints, each near-miss hub satisfies $c-1$,
    and each constraint is the missing one for some near-miss. The question refers to each fact by paraphrase,
    never verbatim. Each hub-affiliation gold names the hub exactly once, never in the first sentence and never
    as ``the sole'' anything, among at least two same-type distractors, and gets seven to ten near-miss
    negatives that never contain the hub name. From $3{,}389$ entities with four or five coined facts each and
    three or four constraints per item, the build yields $15{,}421$ items, with $61{,}732$ hub golds against
    $432{,}044$ hub negatives (${\approx}1{:}7$).
    \item[\textbf{(iv)}] \textbf{Verification:} Six deterministic gates run first: (a)~the fact is searchable in
    the gold; (b)~the entity is present in the gold; (c)~both fall inside one ${\leq}300$-character window;
    (d)~the answer is not in the question; (e)~the answer is not a value in the seed's output column; (f)~the
    constraint intersection is a singleton. Five model-based gates follow: gold faithfulness, a no-document
    parametric probe, a mask test (with the gold removed, the fact must not be recoverable from the item's
    negatives), negative-reader denoising (drop any negative that answers the question), and premise
    plausibility.
\end{itemize}

\begin{table}[h]
\caption{The twelve hard-negative categories and the skill each one teaches.}
\label{tab:sql:negatives}
\begin{center}
\resizebox{0.8\textwidth}{!}{
\begin{tabular}{lll}
\toprule
\textbf{Category} & \textbf{Flips} & \textbf{Skill taught} \\
\midrule
\texttt{same\_entity\_no\_answer}        & omits the fact              & do not accept partial evidence \\
\texttt{wrong\_event}                    & the event                  & event discrimination \\
\texttt{wrong\_date}                     & the date                   & temporal pinning \\
\texttt{wrong\_location}                 & the place                  & spatial pinning \\
\texttt{wrong\_version\_or\_edition}     & a superseded value          & read, do not skim \\
\texttt{wrong\_category\_or\_class}      & the class                  & type discrimination \\
\texttt{right\_attribute\_wrong\_entity} & the entity                 & attribute--entity binding \\
\texttt{namesake\_or\_misspelling}       & the referent               & homonym resistance \\
\texttt{same\_entity\_partial\_evidence} & sufficiency                & stopping criterion \\
\texttt{near\_dup\_missing\_key\_fact}   & the decisive span          & span-level reading \\
\texttt{topically\_adjacent}             & relevance                  & topic vs.\ answer \\
\texttt{lexical\_overlap}                & the referent               & lexical robustness \\
\bottomrule
\end{tabular}}
\end{center}
\end{table}

\textbf{Anti-shortcut pass (\dvfive{}).} Primary-split bridges can be famous real entities (e.g.\ \emph{Google
Photos} in Example~2), which invites early lock-on and distractor capture: committing to the first plausible
candidate, or to a retrieved near-miss, before every constraint is checked. Both failure modes are documented
for long-horizon search agents~\citep{sinha2026illusion,gao2025asearcher} and for LLMs reading noisy retrieved
text~\citep{shi2023distracted,cuconasu2024power,wang2025retrievalaugmented,pham2026sealqa}, and a salient parametric prior makes lock-on
worse~\citep{xie2024knowledgeconflicts}. The pass keeps the seed SQL and refabricates the web puzzle with
DeepSeek-V4-Flash~\citep{deepseekai2026deepseekv4}, which also samples the trajectories: a parametric-prior decoy
receives the longest and most authoritative document in the item (salience inversion), and a real homonym of the
bridge and at least two namesake or misspelling negatives are added.

\subsection{Building \stsql{}: the anchor is a \texttt{WHERE}-clause literal}
\label{app:sql:anchor}

\dstsql{} is built from the synthetic split only, since few primary-split seeds pass the anchor gates. Gold
documents, typed negatives, and verification follow steps (ii) and (iv) of Appendix~\ref{app:sql:sql2s};
Example~3 in Appendix~\ref{app:sql:examples} walks through one item.

\begin{itemize}
    \item[\textbf{(i)}] \textbf{Anchor Selection:} Candidates are string literals under \texttt{EQ},
    \texttt{IN}, or \texttt{LIKE} nodes of the gold query's syntax tree, on a name-bearing column (e.g.\ name,
    title, team, city, product) and outside an enum stop-set such as \{active, yes, male\}. The execution check
    replaces the predicate with a tautology and requires the row set to change, which rejects co-filter decoys
    whose rows a join key or threshold already pins; the membership gate requires the anchor to appear in the
    question but not in the output. Execution must also be non-degenerate ($0<|\text{rows}|<|\text{table}|$),
    and the anchor must support at least three clues, which rejects famous anchors that a single hop resolves
    (e.g.\ \emph{Canada}). A namability judge then rejects five recurring shapes: class labels
    (\texttt{Sedan}), status values (\texttt{in progress}), generic metrics and awards (\texttt{Golden Globe}),
    common personal names (\texttt{Smith}), and ambiguous acronyms or over-coarse places (\texttt{USA}). Anchors
    are not filtered for rarity, since difficulty comes from the clues and decoys. Seeds are instead chosen to
    cover every \emph{SQL group}, the gold queries sharing one structural signature (table-count bucket, CTE
    depth, window-function class, and aggregate class, e.g.\
    \texttt{T5+\textbar{}CTE1\textbar{}WIN:rank\textbar{}AGG:none}); the seed pool keeps all $2{,}518$ groups.
    \item[\textbf{(ii)}] \textbf{Relation Fan-out \& Hub Construction:} Each clue ties the anchor to a coined
    entity through one of $31$ relations over seven anchor types (Table~\ref{tab:sql:relations}). Coined names
    are also checked against the names already in the corpus, and a name that still collides gets a distinctive
    suffix. The anchor is the only entity that satisfies every clue, and each coined near-miss satisfies all but
    one, so an item carries about $43$ hard negatives. The final question lists the clues first and the seed
    question last, with the anchor hidden; otherwise the model answers with the entity name instead of the result
    table.
\end{itemize}

\begin{table}[h]
\caption{The relation menu for \stsql{} clues: $31$ relations over seven anchor types.}
\label{tab:sql:relations}
\begin{center}
\small
\begin{tabular}{lp{0.74\textwidth}}
\toprule
\textbf{Anchor type} & \textbf{Relations} \\
\midrule
place ($10$)         & is the birthplace of $\cdot$ is the hometown of $\cdot$ is the headquarters city of $\cdot$ is the home ground of $\cdot$ is the seat of $\cdot$ hosts the annual ceremony of $\cdot$ is the administrative district of $\cdot$ is the region served by $\cdot$ is the primary service area of $\cdot$ is the coverage territory of \\
organization ($5$)   & administers $\cdot$ develops $\cdot$ is chaired by $\cdot$ is the title sponsor of $\cdot$ was profiled in a feature by \\
person ($5$)         & coached $\cdot$ founded $\cdot$ was the inaugural recipient of $\cdot$ is the lead designer of $\cdot$ was profiled in a feature by \\
institution ($3$)    & is the alma mater of $\cdot$ confers $\cdot$ developed \\
creative work ($3$)  & won $\cdot$ was directed by $\cdot$ was produced by \\
product ($2$)        & is developed by $\cdot$ was designed by \\
program ($3$)        & is administered by $\cdot$ is led by $\cdot$ won \\
\bottomrule
\end{tabular}
\end{center}
\end{table}

\subsection{Corpus and ingestion}
\label{app:sql:corpus}

All fabricated documents are inserted into the retrieval service that the agent queries at evaluation, alongside
real encyclopedic and web text. \dmc{} contributes $1{,}548{,}257$ documents (hop-one clues and hubs), \dvfive{}
${\sim}250$k, and \dstsql{} $591{,}664$.

\clearpage
\subsection{Worked examples}
\label{app:sql:examples}

Three shipped items, one per subset. Titles and quoted text are copied from the item records and the
shipped trajectories; ``\ldots'' marks elisions, and each trajectory is condensed to its decisive steps.

\providecommand{\exlab}[1]{\textcolor{blue!50!black}{\textbf{#1}}}
\newtcolorbox{sqlexample}{colback=black!2,colframe=black!35,boxrule=0.5pt,arc=3pt,
  left=7pt,right=7pt,top=6pt,bottom=6pt,fontupper=\small,before skip=4pt,after skip=10pt}

\begin{sqlexample}
\textbf{Example 1: \dmc{}} (\sqlts{}, synthetic split)\\[1pt]
{\footnotesize Database \texttt{energy\_asset\_management\_and\_capacity\_planning}. Answer: \textbf{Vasewihu Industries}.}

\smallskip
\exlab{Coined metric.} \emph{Average Operational Capacity Score (AOCS)}: ``the average current operational
capacity of non-decommissioned assets, grouped by region and technology''. It is defined in the question's
domain-knowledge block and served by the knowledge tool.

\exlab{SQL leg.} ``Among the non-decommissioned assets with average capacity exceeding 10\,MW, which technology
has the highest Average Operational Capacity Score?''
{\footnotesize
\begin{verbatim}
SELECT t.technology_name, r.region_name,
       AVG(a.current_capacity) AS aocs
FROM assets a
  JOIN technologies t ON a.technology_id = t.technology_id
  JOIN regions r ON a.region_id = r.region_id
WHERE a.decommission_date IS NULL
GROUP BY t.technology_id, r.region_id
\end{verbatim}
}
\noindent$\Rightarrow$ bridge $B$ = \textbf{Gas Supply}, an output cell that the question never names.

\exlab{Hop one: coined facts about $B$.} $B$ has four coined facts; this item uses three. Each has one gold
document and 8--10 typed negatives.

\begin{center}\footnotesize
\begin{tabular}{@{}>{\raggedright\arraybackslash}p{0.19\linewidth}>{\raggedright\arraybackslash}p{0.33\linewidth}>{\raggedright\arraybackslash}p{0.38\linewidth}@{}}
\toprule
\textbf{Fact} & \textbf{Gold document} & \textbf{Sample negatives} \\
\midrule
F1 \emph{Kavedi Initiative} (award) & Gas Supply System Recognized with Kavedi Initiative Award in 2025 Annual
Reliability Assessment & \texttt{wrong\_date}: \ldots{} Passes 2023 Reliability Review with Version 3.0 Upgrade;
\texttt{right\_\allowbreak attribute\_\allowbreak wrong\_\allowbreak entity}: Marlowe Gas System Awarded 2025 Reliability Excellence Prize \\
F2 \emph{Kesebu Cone} (person) & Modular Gas Supply Unit Deployed in Arctic Regions in 2022: Technical
Overview and Leadership & \texttt{wrong\_location}: \ldots{} Tested in Siberian Permafrost Zones \ldots;
\texttt{wrong\_\allowbreak category\_\allowbreak or\_\allowbreak class}: Portable Gas Generator Unit Selected for Remote Arctic Research Stations \\
F4 \emph{Koneri Dynamics} (company) & Koneri Dynamics Completes Acquisition of Gas Supply Network in Europe
Amid Major Infrastructure Consolidation & \texttt{near\_\allowbreak dup\_\allowbreak missing\_\allowbreak key\_\allowbreak fact}: Major Infrastructure Shifts in
Europe Highlight Gas Supply's Role in Energy Transition \\
\bottomrule
\end{tabular}
\end{center}

\exlab{Hop two: the hub.} One hub is minted per three-fact subset, four in all. The gold hub for this item,
\textbf{Vasewihu Industries}, funds F1, retains F2 as an advisor, and co-invests with F4. Each relation has one
gold document, e.g.\ ``Vasewihu Industries Supports Innovation Through Kavedi Initiative Grant'', which also
names two minted distractors, \emph{Bimifidu Labs} and \emph{Bovibiwa Ventures}. The other three hubs are this
item's near-misses, each missing one constraint: \emph{Fawuwuza} (not F4), \emph{Jufowoto Jigi} (not F2), and
\emph{Norenivu Gazette} (not F1). Each relation also has 7--8 hub negatives, e.g.\ \texttt{namesake\_\allowbreak or\_\allowbreak misspelling}:
``Kavedi Initiative Awards Grant to Vasewihu Group in 2021''.

\exlab{Question after the SQL part.} ``For that entity, identify each of the following: (a) the designation
awarded to that entity system that passed the 2025 annual reliability assessment with version 3.2 (b) the lead
developer of the modular that entity unit deployed in arctic regions in 2022 (c) the company that acquired that
entity network in europe during the 2022 major infrastructure consolidation. Then name the single organization
that: funds the entity in (a), retains as an advisor the entity in (b), co-invests with the entity in (c).''

\exlab{Generated Trajectory} (7 searches, 7 pages). (1)~SQL computes AOCS $\Rightarrow$ \emph{Gas Supply}. (2)~One search
per fact. For F1 the top hit is the near-duplicate negative ``\ldots{} 2025 Reliability Assessment: Key Findings
and Performance Metrics'', and the agent opens the gold award page instead; it resolves \emph{Kavedi Initiative},
\emph{Kesebu Cone}, and \emph{Koneri Dynamics}. (3)~One search per hub relation; the agent opens the three gold
affiliation pages and one hub negative, and a combined query confirms the hub. Answer: \textbf{Vasewihu Industries}~\ding{51}.
\end{sqlexample}

\begin{sqlexample}
\textbf{Example 2: \dvfive{}} (\sqlts{}, primary split)\\[1pt]
{\footnotesize BIRD database \texttt{app\_store}. Answer: \textbf{Jivoka Networks}.}

\smallskip
\exlab{Coined metric.} \emph{Photography Subjectivity Score (PSS)}: ``the sum of the Sentiment\_Subjectivity
values across all its user reviews''. It is defined only in the question's domain-knowledge block, since
real databases have no knowledge base.

\exlab{SQL leg.} ``Which Photography genre app has the highest Photography Subjectivity Score (PSS)?''
{\footnotesize
\begin{verbatim}
SELECT App, SUM(CAST(Sentiment_Subjectivity AS REAL)) AS PSS
FROM user_reviews
WHERE App IN (SELECT App FROM playstore
              WHERE Genres = 'Photography')
GROUP BY App ORDER BY PSS DESC LIMIT 20
\end{verbatim}
}
\noindent$\Rightarrow$ bridge $B$ = \textbf{Google Photos} (PSS $95.9$). $B$ is a famous real app, so the
agent's prior about its parent company is exactly what the puzzle must override.

\exlab{Web puzzle.} ``Then, for that entity, according to the 2025 edition of the Mobile App Industry Report,
which company or organization is listed as the parent entity of the Photography-genre app that ranks first by
total sentiment subjectivity? Name only the company or organization.'' The answer is a coined company.

\exlab{Documents the agent retrieved.} The pinning constraint is the source: only the gold document attributes a
parent to the 2025 edition of that report.

\begin{center}\footnotesize
\begin{tabular}{@{}>{\raggedright\arraybackslash}p{0.17\linewidth}>{\raggedright\arraybackslash}p{0.39\linewidth}>{\raggedright\arraybackslash}p{0.34\linewidth}@{}}
\toprule
\textbf{Role} & \textbf{Title} & \textbf{What it says} \\
\midrule
gold \ding{51} & Google Photos: A Deep Dive into Its Features, Privacy, and Market Standing & ``According to the 2025
edition of the Mobile App Industry Report, Google Photos is a subsidiary of Jivoka Networks'' \\
famous prior & Alphabet Inc.\ Annual Report 2025: The Mobile App Industry Report and the Dominance of Google Photos
& Alphabet as parent, per Alphabet's own annual report \\
wrong edition & European Mobile App Industry Report 2025: Photography Apps and Sentiment Leadership & PixelWorks
Ltd., holder of the app's European distribution rights \\
unsourced parent & Google Photos: A Comprehensive Guide to the Photography App and Its Parent Company & PixelWorks
Inc., not attributed to the report \\
contradiction & Google Photos: The Photography App That Isn't a Subsidiary of Any Tech Giant & ``the report does not
list any parent entity'' \\
withholds & Top Photography Apps by Sentiment Subjectivity: 2025 Mobile App Industry Report Analysis & ``the report
does not disclose the parent entity'' \\
namesake & Google Photots: The Photography App That Tops Sentiment Rankings in 2025 & a misspelled app name \\
same attribute & Instagram, Snapchat, Snapseed, VSCO: ``\ldots{} Highest Sentiment Subjectivity'' & a different app \\
wrong year & Mobile App Industry Report 2024, 2019 & earlier editions \\
\bottomrule
\end{tabular}
\end{center}

\exlab{Generated Trajectory} (5 searches, 7 pages). (1)~Three SQL calls list the schema and the photography apps, then
compute PSS $\Rightarrow$ \emph{Google Photos}. (2)~The first two searches return the gold page among the
decoys above. The agent opens the gold page, the Alphabet report, the European edition, the guide, and the
contradicting and withholding pages, and keeps the one claim attributed to the 2025 edition. (3)~Targeted
queries naming \emph{Jivoka Networks}, \emph{PixelWorks}, and \emph{Alphabet} confirm that no other document
makes that attribution. Answer: \textbf{Jivoka Networks}~\ding{51}.
\end{sqlexample}

\begin{sqlexample}
\textbf{Example 3: \dstsql{}} (\stsql{}, synthetic split)\\[1pt]
{\footnotesize Database \texttt{election\_results\_and\_voting\_data\_management}. Answer: the seed query's result table.}

\smallskip
\exlab{Seed.} ``Could you provide the names of the precincts in Wilson county during the 2023-2024 election cycle
that had the highest voter turnout, along with the total number of votes and the most commonly used voting method
in those precincts?'' The gold query, abridged:
{\footnotesize
\begin{verbatim}
WITH PrecinctDetails AS (
  SELECT p.precinct_id, p.precinct_name, e.election_cycle,
         SUM(v.votes) AS total_votes
  FROM precincts p JOIN counties c ON p.county_id = c.county_id
       JOIN votes v ON p.precinct_id = v.precinct_id
       JOIN elections e ON v.election_id = e.election_id
  WHERE c.county_name = 'Wilson' AND e.election_cycle = '2023-2024'
  GROUP BY p.precinct_id, p.precinct_name, e.election_cycle),
RankByTurnout AS (... ROW_NUMBER() OVER
    (ORDER BY total_votes DESC) AS turnout_rank ...),
TopMethods AS (... ROW_NUMBER() OVER (PARTITION BY v.precinct_id
    ORDER BY SUM(v.votes) DESC) AS method_rank ...)
SELECT ... WHERE rb.turnout_rank = 1 AND tm.method_rank = 1;
\end{verbatim}
}
\noindent\exlab{Anchor.} $A$ = \texttt{'Wilson'}: it is in the \texttt{WHERE} clause and in the question, and
not in the output (precinct, vote total, voting method).

\exlab{Hop one: coined clues about $A$.} Each clue has one gold document:

\begin{center}\footnotesize
\begin{tabular}{@{}>{\raggedright\arraybackslash}p{0.21\linewidth}>{\raggedright\arraybackslash}p{0.34\linewidth}>{\raggedright\arraybackslash}p{0.35\linewidth}@{}}
\toprule
\textbf{Clue (coined entity)} & \textbf{Descriptor in the question} & \textbf{Gold document} \\
\midrule
C1 \emph{is the region served by} \emph{Votixen Solutions} & election-technology firm for voter engagement
analytics and digital polling infrastructure, early 2020s & Wilson County's Role in Election Infrastructure and
Voter Engagement \\
C2 \emph{is the primary service area of} \emph{Votixa} & election-technology firm for real-time turnout analytics
and digital ballot verification & Wilson County's Electoral Infrastructure and Local Voting Management \\
C3 \emph{is the coverage territory of} \emph{Votisyn Labs} & election-technology research institute analyzing
voting patterns across U.S.\ counties & Wilson County Election Administration and Voter Engagement in 2023-2024 \\
\bottomrule
\end{tabular}
\end{center}

\exlab{Hop two: near-miss places.} Wilson is the only place satisfying all three clues. Each coined near-miss
satisfies two, with its own affiliation documents: \emph{Tiganefu} (C2, C3), \emph{Fefecuga} (C1, C3), and
\emph{Funusosu} (C1, C2), e.g.\ ``Local Governance and Voter Infrastructure in Wilson County's 2023--2024
Election Cycle'' for Funusosu. Typed negatives attach each clue entity to coined wrong places (\emph{Gofuguje},
\emph{Wigamape}): \texttt{wrong\_relation} ``Gofuguje Announces Expansion of Digital Infrastructure with Votixen
Solutions Partnership''; \texttt{wrong\_date} ``Votixen Solutions Launches 2018 Initiative in Wigamape for Rural
Connectivity''; \texttt{wrong\_\allowbreak version\_\allowbreak or\_\allowbreak edition} ``Votixen Solutions Releases Version 3.2 for Gofuguje's
Municipal Systems''; \texttt{same\_\allowbreak clue\_\allowbreak no\_\allowbreak link} ``Why Wigamape Is Emerging as a Hub for Biotech Research and
Development''.

\exlab{Final question.} ``A single place or settlement satisfies ALL of the following clues: is the region served
by [C1]; is the primary service area of [C2]; is the coverage territory of [C3]. First determine that place or
settlement from the clues above. Then, treating that place or settlement as the referenced entity (it is the
`county\_name' value in the query), answer the original question: Could you provide the names of the precincts in
that place or settlement county during the 2023-2024 election cycle that had the highest voter turnout, \ldots?''

\exlab{Generated Trajectory} (6 searches, 7 pages). (1)~Three searches, one per clue descriptor, surface off-target
election pages. (2)~The agent opens Wilson County pages, including the near-miss pages for \emph{Funusosu} and
\emph{Tiganefu}, each of which matches only two clues. (3)~Searches pairing each coined name with Wilson County
reach all three gold pages. (4)~It runs its own version of the seed query with \texttt{county\_name = 'Wilson'},
which returns the gold table: \emph{Precinct 101}, $220$ votes, \emph{Early Voting}~\ding{51}.
\end{sqlexample}

\section{Search-Originated Construction Details}
\label{app:searchorig}

\subsection{Seed pool and construction}
\label{app:searchorig:seeds}

We collect multi-hop search questions with short verified answers from six public deep-research question sets, listed in Table~\ref{tab:searchorig_seeds}. We remove malformed, duplicate, and unverified QA pairs, leaving 22{,}536 candidate seeds. We require the answer of a seed to be a named entity that a database can hold as a value and that the web can confirm, so we pass every answer through a Google Knowledge Graph gate. Bare numbers, decimals, ordinals, percentages, and yes/no tokens are rejected on sight, four-digit years are kept, and every remaining string is looked up and retained when its top result scores above 50. The gate earns its place in both directions. In S2SQL the answer becomes the anchor that the SQL question names, so it has to be a value the database can key on. In SQL2S the answer stays the final answer of the merged item, where a short unambiguous entity is what keeps grading reliable. This leaves 14{,}233 seeds.

\begin{table}[t]
\centering
\caption{Seed sources.}
\label{tab:searchorig_seeds}
\begin{tabular}{lr}
\toprule
\textbf{Source} & \textbf{Questions} \\
\midrule
REDSearcher (SFT) \citep{chu2026redsearcher} &  9{,}177 \\
OpenSeeker \citep{du2026openseeker} &  8{,}743 \\
DeepDive (RL) \citep{deepdive2025} &  2{,}202 \\
REDSearcher (RL) \citep{chu2026redsearcher} &     979 \\
DeepDive (SFT) \citep{deepdive2025} &     961 \\
WebShaper \citep{webshaper2025} &     474 \\
\midrule
Collected         & 22{,}536 \\
After entity gate & 14{,}233 \\
\bottomrule
\end{tabular}
\end{table}

Each seed is grounded once and then follows a single construction chain. S2SQL reuses the seed's answer as the anchor, while SQL2S mines candidate entities from the search question and retains one that passes the same Knowledge Graph gate. To further scale SQL2S, we add three additional question sets of the same kind: Orbit~\citep{thakur2026orbit}, ToolMind~\citep{yang2025toolmind}, and LoHoSearch~\citep{zhao2026lohosearch}, yielding 24{,}285 additional filtered seeds used only for SQL2S. Table~\ref{tab:searchorig_funnel} reports the remaining data size after each construction stage.

\begin{table}[t]
\centering
\caption{Data size remaining after filtering at each stage of the search-originated construction pipeline.}
\label{tab:searchorig_funnel}
\begin{tabular}{lccc}
\toprule
\textbf{Stage}
& \textbf{SQL2S}
& \textbf{S2SQL}
& \textbf{SQL2S (scale-up)} \\
\midrule
Seeds
& 14{,}233
& 14{,}233
& 24{,}285 \\
Entity grounding
& 13{,}941
& 14{,}233
& 23{,}342 \\
\midrule
Plan
& 13{,}878
& 14{,}232
& 23{,}245 \\
Improve
& 13{,}864
& 14{,}232
& 23{,}229 \\
Build
& 13{,}479
& 14{,}223
& 22{,}502 \\
Bridge
& 13{,}420
& 14{,}193
& 22{,}278 \\
Final improve
& 13{,}420
& 14{,}179
& 22{,}278 \\
\bottomrule
\end{tabular}
\end{table}

\subsection{Shipped databases.}
\label{app:searchorig:dbs}

Each item ships with its own SQLite database. Table~\ref{tab:searchorig_dbs} summarises the 35{,}698 SQL2S databases and the 14{,}179 S2SQL databases. A typical database is a single table with a few hundred peer rows around the bridge; just over a third of the SQL2S databases, and a quarter of the S2SQL databases, join two or more tables.

\begin{table}[t]
\centering
\caption{Shipped SQLite databases for the search-originated arm.}
\label{tab:searchorig_dbs}
\begin{tabular}{lrr}
\toprule
 & \textbf{SQL2S} & \textbf{S2SQL} \\
\midrule
Databases & 35{,}698 & 14{,}179 \\
Tables per database (mean / median) & 1.55 / 1 & 1.36 / 1 \\
Share with $\geq 2$ tables & 37.6\% & 24.8\% \\
Columns per table (mean) & 4.15 & 4.14 \\
Rows per table (mean / median) & 613 / 200 & 348 / 215 \\
Rows per database (mean / median) & 953 / 329 & 473 / 292 \\
\bottomrule
\end{tabular}
\end{table}

\subsection{Worked examples}
\label{app:searchorig:examples}

Two shipped items, one per direction, both from the same construction pipeline in Section~\ref{app:searchorig:seeds}. Titles and quoted text are copied from the item records and the shipped trajectories; ``\ldots'' marks elisions, and each trajectory is condensed to its decisive steps.

\begin{sqlexample}
\textbf{Example 1: SQL2S} (DeepDive SFT seed)\\[1pt]
{\footnotesize Database \texttt{deep\_dive\_sft\_84}. Answer: \textbf{Rob Mayes}.}

\smallskip
\exlab{Seed.} ``A UK comedy show's particular season, concluding twelve years prior to 2020, was helmed by one London-born individual. He also directed a sitcom about civilian surveillance, one episode of which, with `Bust!' in its title, featured an actor born as WWII began. This actor's son connects to a 20th-century playwright, whose drama about personal crisis shares its name with a sci-fi horror sequel. \ldots{} In this film about enhanced sharks, who portrayed the ex-naval commando?'' Gold answer: \textbf{Rob Mayes}.

\exlab{Bridge.} $B$ = \textbf{Bust!}, a title named in the seed's search question and absent from the search answer. It becomes the SQL answer and is stripped from the merged question.

\exlab{SQL leg.} ``Based on the database, which episode achieved the highest quality-adjusted engagement score?'' Hint: that score is $(\texttt{guest\_star\_count}/\texttt{viewership\_millions})\times(\texttt{social\_engagement\_index}\times\texttt{critical\_rating}/100)$. The table \texttt{tv\_episodes} holds 248 rows of same-kind titles (\emph{Lockdown!}, \emph{Breakout!}, \emph{Ambush!}, \ldots).
{\footnotesize
\begin{verbatim}
WITH stage_1 AS (
  SELECT episode_title,
         guest_star_count / NULLIF(viewership_millions, 0) AS ratio
  FROM tv_episodes),
stage_2 AS (
  SELECT s1.episode_title,
         s1.ratio
           * (t.social_engagement_index * t.critical_rating / 100)
           AS quality_score
  FROM stage_1 s1 JOIN tv_episodes t
    ON s1.episode_title = t.episode_title),
ranked AS (
  SELECT episode_title, quality_score,
         RANK() OVER (ORDER BY quality_score DESC) AS rnk
  FROM stage_2)
SELECT episode_title, quality_score, rnk FROM ranked WHERE rnk = 1
\end{verbatim}
}
\noindent$\Rightarrow$ $B$ = \textbf{Bust!} (score $7.62$).

\exlab{Final question.} The seed is kept, except the named episode is replaced by the SQL task: ``\ldots{} one episode of which, based on the database, achieved the highest quality-adjusted engagement score and featured an actor born as WWII began. \ldots{} who portrayed the ex-naval commando?''

\exlab{Generated Trajectory} (21 searches, 7 pages). (1)~SQL lists \texttt{tv\_episodes} and ranks by the hinted formula $\Rightarrow$ \emph{Bust!}. (2)~Queries for a UK sitcom about civilian surveillance open pages for \emph{Spy} (2011) and \emph{Watching}; a later query for the 2008 London-born director of \emph{That Mitchell and Webb Look} reaches \emph{David Kerr}. (3)~The shark-sequel clue opens the \emph{Deep Blue Sea 2} cast list. Answer: \textbf{Rob Mayes}~\ding{51}.
\end{sqlexample}

\begin{sqlexample}
\textbf{Example 2: S2SQL} (OpenSeeker seed)\\[1pt]
{\footnotesize Database \texttt{open\_seeker\_109}. Answer: the gold SQL result table.}

\smallskip
\exlab{Seed.} ``A coastal administrative division of a transcontinental republic, located on the northeastern shoulder of the continent where that republic is the sole European-derived sovereign power, borders a body of water containing a minor archipelago. One constituent of this archipelago bears a name synonymous with a condition of eternal spiritual punishment. This division is also party to a contemporary, unresolved dispute over a segment of a riverine frontier with the political entity lying immediately to its west. \ldots{} Identify this administrative division.'' Gold answer: \textbf{French Guiana}.

\exlab{Bridge.} $B$ = \textbf{French Guiana}. It is named in the SQL question as the department the query filters on, and it must not be recoverable from the result table alone once the question hides it.

\exlab{SQL leg.} ``Based on the database, what was French Guiana's rank by GDP per capita among French overseas departments in 2023, along with its GDP per capita value?'' Hint: GDP per capita (thousands of euros) $=$ GDP (millions of euros) $/$ population (thousands). The table \texttt{overseas\_department\_annual\_stats} has 312 rows over 26 labeled departments (Guadeloupe, Martinique, R\'{e}union, \ldots).
{\footnotesize
\begin{verbatim}
WITH gdp_per_capita AS (
  SELECT department, year,
         gdp_millions_eur / population_thousands
           AS gdp_per_capita_thousands
  FROM overseas_department_annual_stats
  WHERE year = 2023),
ranked_departments AS (
  SELECT department, gdp_per_capita_thousands,
         RANK() OVER (ORDER BY gdp_per_capita_thousands DESC)
           AS rank
  FROM gdp_per_capita)
SELECT department, ROUND(gdp_per_capita_thousands, 4), rank
FROM ranked_departments
WHERE department = 'French Guiana'
\end{verbatim}
}
\noindent$\Rightarrow$ one row: \textbf{French Guiana}, $30.0$, rank $2$.

\exlab{Final question.} The year and the department name are gone; the seed's clues replace them: ``Identify the coastal administrative division \ldots{} bordering a body of water containing an archipelago with a constituent named after eternal spiritual punishment, and involved in an unresolved riverine frontier dispute with the political entity to its west. Based on the database, what was this division's rank by GDP per capita among French overseas departments in 2023, along with its GDP per capita value?''

\exlab{Generated Trajectory} (7 searches, 4 pages). (1)~SQL lists the schema and the 26 department labels; none of the clues is in the table. (2)~A search for a French overseas department with a western river dispute opens the France--Suriname border page; a follow-up on the coastal archipelago confirms \emph{\^Ile du Diable} (Devil's Island) in French Guiana. (3)~The agent reruns the ranking with \texttt{department = 'French Guiana'}. Answer: \textbf{French Guiana}, $30.0$, rank $2$~\ding{51}.
\end{sqlexample}

\section{SFT Data Ablation}
\label{app:sft_ablation}

Table~\ref{tab:sqlorig:ablation} adds each data family in turn, starting from hydra and the nine raw
text-to-SQL sources. Search-originated data raises overall pass@8 from $47.50$ to $54.17$, and SQL-originated
data raises it further to $59.17$, with most of the second gain on \stsql{} ($52.50\rightarrow65.00$).

\begin{table}[t]
\centering\small
\caption{SFT data ablation on Qwen3.5-4B (\%; setup in \S\ref{sec:eval}). Top: the base model without SFT.
Bottom: data families added in turn: hydra and the nine raw text-to-SQL sources (Table~\ref{tab:sqlorig:base}),
then search-originated (Appendix~\ref{app:searchorig}) and SQL-originated data (Table~\ref{tab:sql:shipped}); the
last row is the full recipe.}
\label{tab:sqlorig:ablation}
\begin{tabular*}{\textwidth}{@{\extracolsep{\fill}}lcccccccc@{}}
\toprule
& \multicolumn{4}{c}{\textbf{Pass@8}} & \multicolumn{4}{c}{\textbf{Avg@8}} \\
\cmidrule(lr){2-5}\cmidrule(l){6-9}
\textbf{Training data} & \textbf{Overall} & \textbf{S2SQL} & \textbf{SQL2S} & \textbf{Parallel} & \textbf{Overall} & \textbf{S2SQL} & \textbf{SQL2S} & \textbf{Parallel} \\
\midrule
base (no SFT)        & $30.83$ & $22.50$ & $20.00$ & $50.00$ & $7.71$ & $5.00$ & $3.44$ & $14.69$ \\
hydra $+$ SQL        & $47.50$ & $45.00$ & $22.50$ & $75.00$ & $13.96$ & $11.25$ & $6.25$ & $24.38$ \\
\quad $+$ search-orig.   & $54.17$ & $52.50$ & $30.00$ & $\mathbf{80.00}$ & $20.10$ & $18.75$ & $\mathbf{9.69}$ & $31.88$ \\
\qquad $+$ SQL-orig.      & $\mathbf{59.17}$ & $\mathbf{65.00}$ & $\mathbf{35.00}$ & $77.50$ & $\mathbf{23.75}$ & $\mathbf{26.88}$ & $8.75$ & $\mathbf{35.63}$ \\
\bottomrule
\end{tabular*}
\end{table}

\section{Local Corpus}
\label{sec:app:local_corpus}
Generating teacher trajectories and RL rollouts against a search API is costly at our scale, and results drift as the web changes.
We therefore serve the agent's web tools from a local corpus of 21M documents, which combines real web text from the June 2026 English Wikipedia dump\footnote{\url{https://dumps.wikimedia.org/enwiki/20260601/}} and a subset of the CC-MAIN-2025-26 snapshot of FineWeb~\citep{penedo2024the}, with the fabricated gold documents and hard negatives of Sec~\ref{sec:data:sqlorig} (ingestion details in Appendix~\ref{app:sql:corpus}).
Our design goal is to reproduce the observations a search API returns, so that the agent's tool interface is unchanged when it moves to live search.
In particular, such APIs return query-dependent snippets: each result shows the part of the page that best matches the query rather than a fixed summary.
To support it, we split each document into chunks by greedily packing whole sentences up to 500 characters, without overlap.
Both documents and chunks are indexed for hybrid retrieval, which combines BM25~\citep{bm25} with dense cosine similarity. 
\texttt{web\_search} retrieves in two stages. 
Document-level retrieval first selects 100 candidate documents. 
At the chunk level, the same hybrid index then retrieves a pool of chunks from these candidates, and a cross-encoder reranker (Qwen3-Reranker-0.6B; \citealp{qwen3embedding2025}) reorders the pool. 
Each document is ranked by its highest-scoring chunk (MaxP; \citealp{dai2019deeper}), and the top $k \leq 20$ documents are returned, one result per page, each with its title, URL, date, and a snippet taken from that chunk.
\texttt{visit\_web} returns the full text of a document, but only for URLs that appeared in an earlier search result, so the agent cannot fetch pages it has not found.

\section{Hybrid Agent Harness}
\label{app:harness}

A question is only hybrid if solving it requires the agent to obtain evidence from both structured and unstructured sources. Our harness provides the common interaction protocol through which the agent accesses these sources. It is a native tool-calling loop: the model takes a turn, the requested tools are executed, their observations are appended to the context, and the process repeats until the agent commits an answer or exhausts its budget. Trajectories are kept in the model's native tool-calling format rather than encoded into prompt text and parsed back out. The protocol seen during training is therefore the same one used when the model is served, and each rollout can be judged, replayed, and packed for supervised fine-tuning without format conversion. We use this harness for teacher trajectory collection, reinforcement-learning rollouts, and evaluation of our trained checkpoints.

\subsection{Tool Interface}
\label{app:harness:tools}

The agent is given neither the database schema nor any web document in its prompt, so both sources must be accessed through the tools in Table~\ref{tab:harness_tools}. On the database side, \texttt{run\_sql} is the only interface to the underlying SQLite database and supports both schema discovery and subsequent query execution. Three knowledge-lookup tools expose database-specific metrics and their definitions when needed. On the web side, \texttt{web\_search} returns ranked results with short snippets, while \texttt{visit\_web} retrieves the full text of a selected document. The same interface supports both SQL2S and S2SQL: in SQL2S, a SQL result provides the bridge entity for subsequent web research, whereas in S2SQL, an entity resolved from web evidence is used in the subsequent SQL query. The harness does not prescribe either ordering; the model selects its next tool from the same tool set at every turn.

\begin{table}[h]
\centering
\small
\caption{Tools exposed by the hybrid agent harness.}
\label{tab:harness_tools}
\begin{tabular}{@{}ll@{}}
\toprule
Tool & What it does \\
\midrule
\multicolumn{2}{@{}l}{\emph{Web}} \\
\texttt{web\_search(query, k)} & Returns up to $k \le 20$ ranked search results with short snippets \\
\texttt{visit\_web(url)} & Returns the full text of a selected web document \\
\addlinespace
\multicolumn{2}{@{}l}{\emph{Database}} \\
\texttt{run\_sql(sql)} & Executes a read-only SQL query \\
Knowledge lookup (3 tools) & Lists database-specific metrics and returns their definitions \\
\addlinespace
\multicolumn{2}{@{}l}{\emph{Answer}} \\
\texttt{final\_answer(answer)} & Submits the answer and terminates the episode \\
\bottomrule
\end{tabular}
\end{table}

\subsection{Episode Execution}
\label{app:harness:execution}
Tool failures are returned to the model as ordinary observations rather than terminating the episode, allowing the agent to revise an invalid call or try a different action. An episode ends when \texttt{final\_answer} succeeds, when a model turn contains no tool call, or when the turn or wall-clock budget is exhausted. As the remaining budget shrinks, the harness injects a short reminder to stop exploring and commit an answer, specialized to the expected output type. For S2SQL, the agent submits an executable SQL query; for SQL2S and other text-answer tasks, it submits a concise textual answer.

\subsection{Agent Harness Comparison}
\label{app:harness:comparison}

We compare our hybrid agent harness with two open-source agent frameworks, smolagents~\citep{smolagents2024} and MiroFlow~\citep{miroflow2025}, while holding the backbone model fixed. All systems are evaluated on the HybridDeepResearch~\citep{hdrbench2026} hard subset under the same underlying SQL and retrieval environment. As shown in Table~\ref{tab:harness_comparison}, our harness achieves the highest overall pass@8 for all three evaluated backbones. The improvement is most pronounced for Qwen3.5-397B-A17B-FP8, while the gap is smaller for GLM-5.2.

\begin{table*}[t]
\centering
\caption{Agent harness comparison on HybridDeepResearch (pass@8, \%). Backbones: Qwen3.5-4B,
Qwen3.5-397B-A17B-FP8 (Qwen3.5-397B), and GLM-5.2-NVFP4 (GLM-5.2).}
\label{tab:harness_comparison}
\resizebox{\textwidth}{!}{
\begin{tabular}{lcccccccccccc}
\toprule
& \multicolumn{4}{c}{\textbf{Hybrid Agent (Ours)}} 
& \multicolumn{4}{c}{\textbf{smolagents}}
& \multicolumn{4}{c}{\textbf{MiroFlow}} \\
\cmidrule(lr){2-5}
\cmidrule(lr){6-9}
\cmidrule(lr){10-13}
\textbf{Model}
& \textbf{Overall} & \textbf{S2SQL} & \textbf{SQL2S} & \textbf{Parallel}
& \textbf{Overall} & \textbf{S2SQL} & \textbf{SQL2S} & \textbf{Parallel}
& \textbf{Overall} & \textbf{S2SQL} & \textbf{SQL2S} & \textbf{Parallel} \\
\midrule

Qwen3.5-4B
& \textbf{30.83} & 22.50 & 20.00 & 50.00
& 25.83 & 12.50 & 7.50 & 57.50
& 28.33 & 15.00 & 17.50 & 52.50 \\

Qwen3.5-397B
& \textbf{52.50} & 52.50 & 32.50 & 72.50
& 45.83 & 47.50 & 22.50 & 67.50
& 44.17 & 45.00 & 30.00 & 57.50 \\

GLM-5.2
& \textbf{63.33} & 65.00 & 47.50 & 77.50
& 62.50 & 55.00 & 52.50 & 80.00
& 60.83 & 55.00 & 55.00 & 72.50 \\

\bottomrule
\end{tabular}
}
\end{table*}

\section{Evaluation Details}
\label{app:eval}

This section details the evaluation protocol of \S\ref{sec:eval}: the two search-only benchmarks, the shared inference configuration, Spider~2.0-lite, and the judge prompts. HybridDeepResearch needs nothing beyond \S\ref{sec:eval}; every model is evaluated in the same SQL and retrieval environment.

\subsection{BrowseComp Evaluation}
\label{app:eval:browsecomp}

BrowseComp~\citep{wei2025browsecomp} is run against the live web, and the agent is equipped with five tools:
\begin{itemize}
    \item \texttt{web\_search}: searches the live web;
    \item \texttt{web\_fetch}: retrieves the contents of a selected webpage;
    \item \texttt{pdf\_read}: downloads and extracts text from PDF documents;
    \item \texttt{read\_file}: reads locally materialized files when needed;
    \item \texttt{final\_answer}: submits the final answer and terminates the episode.
\end{itemize}

\paragraph{Web search.}
\texttt{web\_search} first queries the Brave Search API. If Brave returns no results, a query containing multiple quoted spans is rewritten into an OR-style query and issued to Brave again. If the second Brave request also returns no results, the search falls back to Serper (\texttt{google.serper.dev}), which returns Google search results. Tavily is not used in this evaluation.

\paragraph{Webpage retrieval.}
\texttt{web\_fetch} first attempts to retrieve a webpage through Jina Reader (\texttt{r.jina.ai}) when a Jina API key is available. If this fails, the page is fetched directly using a local HTTP client and converted to text with MarkItDown.

PDF documents are handled by \texttt{pdf\_read}. The tool first downloads the PDF and attempts text extraction with \texttt{pypdf}; if this fails, it falls back to Jina Reader.

The live-web environment requires \texttt{BRAVE\_API\_KEY} and \texttt{SERPER\_API\_KEY}; \texttt{JINA\_API\_KEY} is optional. API key values are neither stored nor reported.

\subsection{BrowseComp-Plus Evaluation}
\label{app:eval:browsecomp_plus}

The BrowseComp-Plus~\citep{chen2025browsecompplusfairtransparentevaluation} corpus is indexed with the same hybrid retrieval backend as our local corpus. Retrieval operates directly over full documents; we do not use a separate chunk index or local reranker.

The agent is equipped with three tools:
\begin{itemize}
    \item \texttt{corpus\_search(query)}: performs semantic retrieval over the fixed corpus and returns the top five documents, including \texttt{DOCID}, URL, cosine score, and retrieval snippet;
    \item \texttt{corpus\_fetch(docid)}: retrieves the full text of exactly one document identified by a \texttt{DOCID} returned by \texttt{corpus\_search};
    \item \texttt{final\_answer}: submits the final answer and terminates the episode.
\end{itemize}

\texttt{corpus\_fetch} performs an exact lookup using the returned \texttt{DOCID}; it does not issue another semantic search and does not fetch the external webpage associated with the URL.

\subsection{Inference Configuration}
\label{app:eval:inference}

We use model-specific decoding settings while keeping the interaction budget fixed across model families unless otherwise noted.

\paragraph{Qwen models.}
All Qwen models use a maximum of 300 interaction turns, a maximum of 32,768 generated tokens, and a wall-clock timeout of 3,600 seconds. Decoding uses temperature 0.6, top-$p$ 0.95, and top-$k$ 20. They are
evaluated on 8$\times$ NVIDIA H200 GPUs, except on Spider~2.0-lite (Appendix~\ref{app:eval:spider2lite}).

\paragraph{GLM-5.2.}
GLM-5.2 uses a maximum of 300 interaction turns, a maximum of 65,536 generated tokens, and a wall-clock timeout of 3,600 seconds. We set reasoning effort to \texttt{max}, temperature to 1.0, and top-$p$ to 0.95. We do not explicitly set top-$k$.
GLM-5.2 is evaluated on 8$\times$ NVIDIA B200 GPUs.

\paragraph{Claude Sonnet 4.6.}
Claude Sonnet 4.6 uses a maximum of 300 interaction turns, a maximum of 32,768 generated tokens, and a wall-clock timeout of 3,600 seconds. We set reasoning effort to \texttt{medium}. Other decoding parameters use the model's default settings.

\subsection{Spider~2.0-lite Evaluation}
\label{app:eval:spider2lite}

Spider~2.0-lite~\citep{lei2025spider2} contains 547 instances, of which 135 (\texttt{local}) are posed on SQLite databases distributed with the benchmark. The remaining 412 (\texttt{bq}, \texttt{ga}, \texttt{sf}, and \texttt{sf\_bq}) query cloud data warehouses and can only be executed through BigQuery or the benchmark's hosted Snowflake account. We use all 135 \texttt{local} instances, which span 30 databases, so the subset is determined by database engine alone and not by difficulty or results.

\paragraph{Tools.} The evaluation uses the agent harness of Appendix~\ref{app:harness} with web access removed and, unlike HybridDeepResearch, with the database schema given in the prompt. The agent is equipped with two tools:
  \begin{itemize}
      \item \texttt{run\_sql(sql, description)}: executes one read-only query (\texttt{SELECT}, \texttt{WITH}, \texttt{PRAGMA}, or \texttt{EXPLAIN}) on the instance's SQLite database and returns at most 500 rows, with a 120-second limit per query;
      \item \texttt{final\_answer(answer)}: submits the final answer and terminates the episode.
  \end{itemize}
The system prompt lists only these two tools. A call to any other tool returns an error message to the agent and does not end the episode.

\paragraph{Prompt.} The user message contains the question, the instance's external-knowledge document if it has one, and the full database schema: the \texttt{CREATE TABLE} and \texttt{CREATE VIEW} statements of every table as stored in \texttt{sqlite\_master}, including column types and primary- and foreign-key constraints. No sample rows are included; the agent can inspect values with \texttt{run\_sql}. Thirteen of the 135 instances reference an external-knowledge document (eight distinct documents), which is inserted into the prompt verbatim.

\paragraph{Final answer \& graded query.} The agent must submit, through \texttt{final\_answer}, the single query that answers the question in a fenced \texttt{sql} block. If a submission contains no SQL query, the harness rejects it and asks for the query, up to two times. The check is structural (a fenced \texttt{sql} block, \texttt{SELECT}\,\ldots\,\texttt{FROM}, or \texttt{WITH}\,\ldots\,\texttt{AS (}), so a prose answer that merely contains the word ``select'' or ``with'' does not pass. We grade only the query submitted through \texttt{final\_answer}: the last fenced code block of its answer or, if it has none, the first SQL statement in it (\texttt{SELECT}\,\ldots\,\texttt{FROM} or \texttt{WITH}\,\ldots\,\texttt{AS (}), truncated at its first semicolon. Queries the agent ran with \texttt{run\_sql} are never graded.

\paragraph{Judging.} We use the official Spider~2.0-lite execution-accuracy metric. The submitted query is executed on the instance's database, and its result is compared with the official gold execution results rather than with a gold query, since only 24 of the 135 instances ship one. We port the official comparison (\texttt{compare\_multi\_pandas\_table} in \texttt{spider2-lite/evaluation\_suite/evaluate.py}) line by line. A prediction is correct if it matches at least one of the instance's acceptable gold results.
A prediction matches a gold result if every gold column, restricted to the instance's \texttt{condition\_cols} when these are given, equals some column of the prediction. Row order is ignored, numbers are compared with an absolute tolerance of 0.01, missing values count as 0, and extra predicted columns are allowed. As in the official script, the predicted result is round-tripped through CSV before comparison. A query that fails to execute is scored as incorrect.

\paragraph{Execution limit.} The official evaluator sets no execution limit for SQLite databases. We stop a query after 600 seconds. To check that this limit does not affect any result, every query that reached it was re-executed with a one-hour limit. Four submitted queries reached 600 seconds, two from the base model and two from the RL model; none of them finished within an hour, and they are scored as incorrect.

\paragraph{Inference.} The base model and the hybrid SFT and RL checkpoints are served in bf16 with vLLM~0.28~\citep{vllm} (maximum model length 262{,}144, prefix caching), with data parallelism 4 on four NVIDIA B200 GPUs. Decoding follows Appendix~\ref{app:eval:inference} with thinking mode on. Because each instance requires committing a single query rather than long-horizon research, we use 50 interaction turns and a wall-clock timeout of 1{,}800 seconds per trajectory.

\subsection{LLM Judge Prompts}
\label{app:judge_prompts}

Our HybridDeepResearch judges are adapted from the SimpleQA~\citep{wei2024measuring} judging format, with task-specific modifications for table-valued S2SQL outputs and textual SQL2S/Parallel outputs. BrowseComp and BrowseComp-Plus use the standard BrowseComp-style extracted-answer judging protocol without modification, so we do not reproduce that prompt here.

\newcommand{\promptslot}[1]{\textcolor{violet!80!black}{\{#1\}}}
\newtcblisting{promptlisting}{breakable,listing only,colback=gray!4,colframe=gray!35,boxrule=0.4pt,arc=2pt,
  left=4pt,right=4pt,top=3pt,bottom=3pt,before skip=4pt,after skip=4pt,
  listing options={basicstyle=\scriptsize\ttfamily,columns=fullflexible,keepspaces=true,breaklines=true,
    breakindent=1em,
    emph={CORRECT},emphstyle={\color{green!45!black}\bfseries},
    emph={[2]INCORRECT},emphstyle={[2]\color{red!65!black}\bfseries},
    escapeinside={(*}{*)},
    literate={```}{{\color{gray!75}\textasciigrave\textasciigrave\textasciigrave}}3 {—}{{{\rmfamily\textemdash}}}1 {`}{{\textasciigrave}}1 {'}{{\textquotesingle}}1}}

\paragraph{S2SQL table judge.}\leavevmode
\begin{promptlisting}
Your job is to look at a question, a gold SQL result table, and an agent SQL result table, and then assign a grade of either ["CORRECT", "INCORRECT"].

The agent answered a natural-language question by running SQL. Both tables below were produced by re-executing SQL on the same database: the gold table from the verified gold SQL, and the agent table from the agent's SQL. Grade whether the agent table conveys the same answer as the gold table for the question, judged by meaning rather than exact table shape.

Judge semantic equivalence: do not penalize purely presentational differences.

Also note the following things:
- First identify the key value(s) that answer the question: entities, counts, metrics, and any group or protocol average the question asked for. Judge those. Treat as presentational and do not penalize: column names, JSON key names, extra or missing non-essential columns, table shape, and row order when order was not asked for.
- Row identifiers from different columns or namespaces are presentational when the asked-for metrics match. Different identifier strings alone do not make the table INCORRECT. Different metrics or a different set of entities is INCORRECT.
- Group labels may differ if each group's counts and metric lists still match.
- Numbers match within a small rounding margin. For percentiles and other window-function scores, a difference of about 0.01 or less is CORRECT when the rest of the row matches. Extra or fewer decimals are CORRECT when they round to the gold at the gold's precision, or differ only by that window-function noise. A shifted distribution or a different entity or value is INCORRECT.
- If the question asks for rank: the same entities in the same order are CORRECT even if one table uses a global rank and the other ranks only inside the result set. A different set of entities or a different order is INCORRECT.
- If the question asks for a group or protocol average, or a difference from such an average, that number is a key value. It must match gold's grouping and scope. A differently scoped average is INCORRECT even when the top entities and their raw scores match.
- Grade INCORRECT when the agent table contradicts those key values, is empty or missing while gold is not, or buries the answer among many unrelated rows without isolating it.

Here is a new example. Simply reply with either CORRECT or INCORRECT. Don't apologize or correct yourself if there was a mistake; we are just trying to grade the answer.

```
Question: (*\promptslot{question}*)
Gold table: (*\promptslot{gold\_table}*)
Agent table: (*\promptslot{agent\_table}*)
```

Grade the agent table of this new question as one of:
A: CORRECT
B: INCORRECT

Just return the letters "A" or "B", with no text around it.
\end{promptlisting}

\paragraph{SQL2S and Parallel answer judge.}\leavevmode
\begin{promptlisting}
Your job is to look at a question, a gold target, and a predicted answer, and then assign a grade of either ["CORRECT", "INCORRECT"].

First, I will give examples of each grade, and then you will grade a new example.

The following are examples of CORRECT predicted answers.

```
Question: What are the names of Barack Obama's children?
Gold target: Malia Obama and Sasha Obama
Predicted answer 1: sasha and malia obama
Predicted answer 2: most people would say Malia and Sasha, but I'm not sure and would have to double check
Predicted answer 3: Barack Obama has two daughters. Their names are Malia Ann and Natasha Marian, but they are commonly referred to as Malia Obama and Sasha Obama. Malia was born on July 4, 1998, and Sasha was born on June 10, 2001.
```

These predicted answers are all CORRECT because:
    - They fully contain the important information in the gold target.
    - They do not contain any information that contradicts the gold target.
    - Only semantic meaning of the SAME entity or value matters; capitalization, punctuation, grammar, and order don't matter. A related, nearby, or commonly confused different entity is not a semantic match.
    - Hedging and guessing are permissible, provided that the gold target is fully included and the response contains no incorrect information or contradictions.

```
Question: Which footballer is nicknamed "El Pibe de Oro"?
Gold target: Diego Armando Maradona
Predicted answer: Diego Maradona
```

This predicted answer is CORRECT because it is a shortened form of the same person's name (a commonly omitted middle name). It still identifies the same person as the gold target.

```
Question: What was the name of the UEFA club competition that ran from 1971 until it was rebranded in 2009?
Gold target: UEFA Cup
Predicted answer: UEFA Europa League (formerly the UEFA Cup)
```

This predicted answer is CORRECT because it names the same competition by its current official name and explicitly identifies the gold target as the former name.

```
Question: Which country is the origin of record CB752021?
Gold target: Burundi
Predicted answer: The country is Burundi.

The top destination nations by reliability score from Burundi are:
1. Nicaragua
2. Fiji
3. Macao
4. Algeria

Verification of clues:
- Burundi is a World Bank Group member.
- Hafsa Mossi was a Burundian politician and minister.
```

This predicted answer is also CORRECT because:
    - It clearly commits to the gold target as the answer to the question ("The country is Burundi").
    - Extra supporting detail, explanations, verification, or related lists after the core answer are allowed and should not change the grade, as long as they do not contradict the gold target.

The following are examples of INCORRECT predicted answers.

```
Question: What are the names of Barack Obama's children?
Gold target: Malia and Sasha
Predicted answer 1: Malia.
Predicted answer 2: Malia, Sasha, and Susan.
Predicted answer 3: Barack Obama does not have any children.
Predicted answer 4: I think it's either Malia and Sasha. Or it could be Malia and Jackie. Or it could be Joey and Malia.
Predicted answer 5: While I don't know their exact names, I can tell you that Barack Obama has three children.
Predicted answer 6: It's possible you may mean Betsy and Olivia. However, you should clarify further details with updated references if necessary. Is that the correct answer?
Predicted answer 7: It may be the case that Obama's child is named James. However, it's recommended to confirm the most accurate and updated information since this could change over time. This model may not always reflect the most current information.
Predicted answer 8: I don't know.
Predicted answer 9: I need more context about which Obama you are talking about.
Predicted answer 10: Without researching the web, I cannot answer this question. However, I can tell you that Barack Obama has two children.
Predicted answer 11: Barack Obama has two children. I know that one of them is Malia, but I'm not sure about the other one.
```

These predicted answers are all INCORRECT because:
    - A factual statement in the answer contradicts the gold target. Incorrect statements that have some hedging (e.g., "it is possible that", "although i'm not sure, i think") are also considered incorrect.
    - Or the important information in the gold target is not fully included in the answer (refusal, partial answer, or vague hedge that never commits to the gold target).

```
Question: Which uncontacted people live on North Sentinel Island?
Gold target: Sentinelese
Predicted answer: Jarawa
```

This predicted answer is INCORRECT because Jarawa is a different ethnic group, even though both are indigenous to the Andaman Islands and are often mentioned together.

```
Question: Which New York City borough contains JFK Airport?
Gold target: Queens
Predicted answer: Brooklyn
```

This predicted answer is INCORRECT because Brooklyn is a different, nearby place — not the same borough as the gold target.

Also note the following things:
- Grade based on whether the predicted answer correctly answers the question relative to the gold target. Focus on the committed core answer (often the first clear statement of the entity/value asked for). Do not mark CORRECT answers INCORRECT merely because the response is long or includes extra context, reasoning, verification, or related tables/lists after stating the gold target.
- Extra information is fine when it does not replace or contradict the gold target. Extra information that names a different primary answer, or that contradicts the gold target, makes the prediction INCORRECT.
- For grading questions where the gold target is a number, the numeric value must be correct. Do not accept a nearby but different number.
    - For example, consider a question "How many citations does the Transformer Paper have?" with gold target "120k". Predicted answer "120k" is CORRECT. Predicted answers "124k", "115k", "100k", "113k", "around 120k", and "more than 50k" are INCORRECT.
    - Units stated in the question may be omitted in the predicted answer. For the question "What is the height of Jason Wei in meters?" with gold target "1.73 m", predicted answers "1.73 m" and "1.73" are CORRECT; "1.75" and "1.75 m" are INCORRECT.
    - Extra or fewer decimal places are CORRECT only when they round to the same value as the gold target (or the gold rounds to the prediction at the gold's precision). For gold "1.73", predicted "1.732" is CORRECT; for gold "1.732", predicted "1.73" is CORRECT. A different value after rounding (e.g. gold "1.73" vs predicted "1.75") is INCORRECT.
- The gold target may contain more information than the question. In such cases, the predicted answer only needs to contain the information that is in the question.
    - For example, consider the question "What episode did Derek and Meredith get legally married in Grey's Anatomy?" with gold target "Season 7, Episode 20: White Wedding". Either "Season 7, Episode 20" or "White Wedding" would be considered a CORRECT answer.
- Do not punish predicted answers if they omit information that would be clearly inferred from the question.
    - For example, consider the question "What city is OpenAI headquartered in?" and the gold target "San Francisco, California". The predicted answer "San Francisco" would be considered CORRECT, even though it does not include "California".
    - Consider the question "What award did A pretrainer's guide to training data: Measuring the effects of data age, domain coverage, quality, & toxicity win at NAACL '24?", the gold target is "Outstanding Paper Award". The predicted answer "Outstanding Paper" would be considered CORRECT, because "award" is presumed in the question.
    - For the question "What is the name of Barack Obama's wife?", the gold target is "Michelle Obama". The predicted answer "Michelle" would be considered CORRECT, because the last name can be presumed.
- Do not punish for typos in people's name if it's clearly the same name.
    - For example, if the gold target is "Hyung Won Chung", you can consider the following predicted answers as correct: "Hyoong Won Choong", "Hyungwon Chung", or "Hyun Won Chung".
- Related is not the same. A predicted answer that names a different person, place, ethnic group, organization, work, or event from the gold target is INCORRECT, even if the two are nearby, historically linked, commonly confused, or belong to the same country, family, or category.
- A shortened, reordered, or unofficial form of the SAME proper name is CORRECT (dropping a middle name, particle, or commonly omitted suffix still identifies the same person or entity).
- A former official name, current official name, or well-established alias of the SAME entity is CORRECT. Naming a different entity, a successor that is not the same thing, or a merely related item is INCORRECT.

Here is a new example. Simply reply with either CORRECT or INCORRECT. Don't apologize or correct yourself if there was a mistake; we are just trying to grade the answer.

```
Question: (*\promptslot{question}*)
Gold target: (*\promptslot{target}*)
Predicted answer: (*\promptslot{predicted\_answer}*)
```

Grade the predicted answer of this new question as one of:
A: CORRECT
B: INCORRECT

Just return the letters "A" or "B", with no text around it.
\end{promptlisting}

\section{Failure Analysis of the Base, SFT, and RL Models}
\label{app:failure}

This section details the failure analysis behind \S\ref{sec:analysis:sft_rl}. We analyze all 2,880 rollouts of
the base, SFT, and RL models on the 120-question hard subset (3 models $\times$ 120 questions $\times$ 8 samples). The
three models receive identical prompts, tool schemas, and decoding parameters (Appendix~\ref{app:eval:inference}),
and a rollout that ends without a \texttt{final\_answer} call counts as wrong. Appendix~\ref{app:failure:outcomes}
assigns every rollout to exactly one outcome, and Appendix~\ref{app:failure:uncertain} examines how the agents justify
the answers they submit. Of the 120 questions, 35 are never solved by any of the three models (9 \stsql{},
21 \sqlts{}, and 5 Parallel).

\subsection{Outcome of every rollout}
\label{app:failure:outcomes}

\textbf{Evidence tracing.} We normalize strings by lowercasing them and removing accents and punctuation, and we count
a string as present in a text when it matches on whole words. We trace two strings through every rollout, the
\emph{bridge} and the \emph{gold answer}. For \stsql{}, the bridge is the web-resolved entity (plus its database
spelling when it differs); for \sqlts{}, it is the first cell of the gold SQL result. For Parallel, the SQL candidates
are the rows of the gold SQL result, and the web candidates are the annotated search answers. A string is
\emph{surfaced} if it appears in a \texttt{web\_search} result or a visited page, and \emph{in reasoning} if it appears
in the model's reasoning. On \stsql{}, the model uses the \emph{right entity} if the bridge appears in its final text
or submitted SQL. Paraphrases and entities encoded by ID do not match, so all evidence rates are lower bounds.

\textbf{Outcome categories.} Each rollout receives the first category it matches, in this order.
\stsql{}: correct; no answer; prose answer without SQL; SQL fails to execute; wrong entity, with the right one never
surfaced; wrong entity, with the right one surfaced; right entity with the wrong row count, the wrong columns, or
wrong values.
\sqlts{}: correct; no answer; bridge not retrieved by SQL; gold never surfaced; gold surfaced but never in reasoning;
gold in reasoning but not answered.
Parallel: correct; no answer; gold never in the SQL results; gold never on the web; gold in both sources, with the
answer in the web candidates only, in the SQL candidates only, or in neither.

\begin{figure}[t]
\centering
\includegraphics[width=\linewidth]{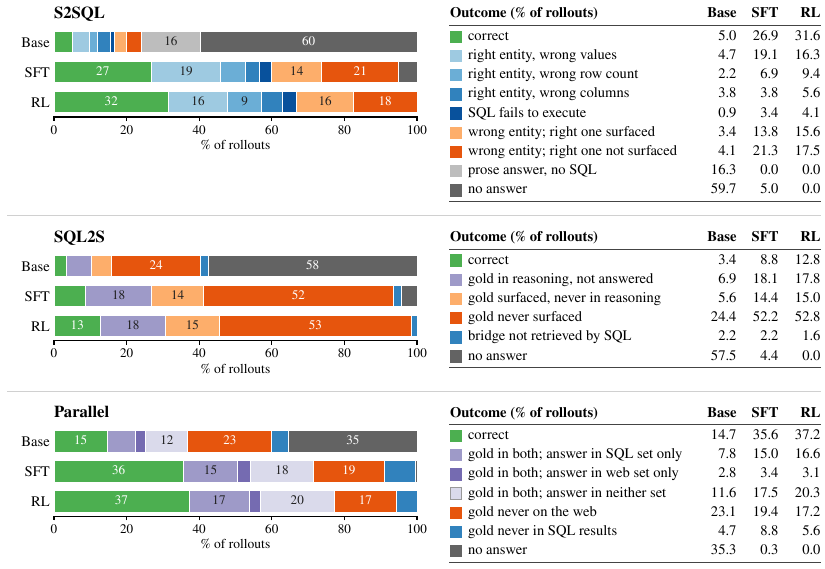}
\caption{Outcome of every rollout, in \% of the 320 rollouts per model and task type (40 questions $\times$ 8
samples). The categories are mutually exclusive and are assigned by the first matching rule listed under
\emph{Outcome categories}; each table lists them in bar order. \emph{Surfaced} means that the string appears in a
search result or a visited page. Bar labels are rounded to whole percentages.}
\label{fig:app:failure}
\end{figure}

\textbf{\stsql{}: wrong entity and wrong SQL.} The base model usually fails before it builds a query: 59.7\% of its
rollouts end without an answer, and 16.3\% answer in prose without SQL. The trained models fail for two reasons of
similar size. The first is resolving the wrong bridge entity (35.0\% of SFT and 33.1\% of RL rollouts), often a
plausible neighbor of the right one, such as Chad, Egypt, or Niger for Sudan, Pakistan for Chile, or Denmark for
Iceland. In 13.8\% and 15.6\% of rollouts, the right entity does appear in a search result or a visited page, but the
query is built on another one. When it appears only in a visited page, it is usually a passing mention in a long page
about something else, such as one row of a table of countries. The second is building wrong SQL around the right entity (29.7\% and 31.3\%), most often with wrong
values. Malformed SQL is rare: at most 4.1\% of rollouts submit a query that fails to execute.

\textbf{\sqlts{}: the search hop.} The SQL hop succeeds: the bridge never appears in the SQL results in only 2.5\% of
SFT and 1.6\% of RL rollouts. The search hop does not. For both trained models, the largest category is a gold
answer that never surfaces (52.2\% of SFT and 52.8\% of RL rollouts). Counting every category, the gold answer never
appears in any search result or visited page in 59.4\% and 54.1\% of rollouts, and accuracy in those rollouts is at
most 2.1\%. When the gold answer does surface, the models often fail to use it: in 14.4\% and 15.0\% of rollouts it
never enters the reasoning, and in 18.1\% and 17.8\% it enters the reasoning but a different answer is submitted.
Typical wrong answers are the bridge itself (about 7\% of wrong \sqlts{} answers), a salient entity next to the bridge
(the capital Moroni for a question whose bridge is the Comoros and whose answer is the village Iconi), and a name
copied from a coined bridge (``Jim Taylor'' for a bridge named Taylortown).

\textbf{Parallel: coverage and the intersection.} In 8.8\% of SFT and 5.6\% of RL rollouts, the answer is wrong and
the gold answer never reaches the SQL results; in a further 19.4\% and 17.2\%, it reaches the SQL results but never
the web. The largest remaining failure is the intersection, which more rollouts reach as coverage improves: the gold
answer appears in both sources, but the submitted answer is in only one candidate set (18.4\% and 19.7\%) or in
neither (17.5\% and 20.3\%). Answers in neither set are often neighbors of the gold entity, such as Tajikistan for
Turkmenistan or Greece for Serbia. When the answer satisfies only one side, that side is SQL far more often than the
web (15.0\% vs.\ 3.4\% for SFT and 16.6\% vs.\ 3.1\% for RL), which suggests that the models trust the structured list
and do not verify the web constraints.

\textbf{Repeated errors and judge noise.} Wrong answers repeat across samples: for the trained models, a single wrong
answer covers at least 4 of 8 samples on up to 18 of 40 \sqlts{} questions and up to 14 of 40 Parallel questions.
Judge noise is negligible: only 0.87\%, 0.29\%, and 0.15\% of wrong base, SFT, and RL rollouts receive a split judge
vote (1--4 of 5 votes correct).

\textbf{Base model.} Half of the base model's rollouts (50.8\%) end without an answer. In 43.2\% of all its rollouts, the
last turn is a single reasoning turn that exhausts the 32K-token limit without calling a tool, and in 412 of these 415
turns, fewer than half of the sentences are unique. In one \sqlts{} rollout, the last turn contains 1,564 sentences,
of which only 38 are distinct.

\subsection{Commits justified by time, effort, or a guess}
\label{app:failure:uncertain}

\textbf{Setup.} For every evaluation rollout that submits an answer (a commit: 472 base, 929 SFT, 960 RL), we flag the commit when its rationale, the last three assistant messages up to and including the \texttt{final\_answer} call, \textbf{(1) cites a time or turn budget}, \textbf{(2) cites the effort already spent}, or \textbf{(3) calls the answer a guess}. The flag is a deterministic, case-insensitive phrase match (Table~\ref{tab:app:markers} lists frequent phrases).

\begin{table}[!ht]
\centering\small
\caption{Frequent phrases that flag a commit, per family, in SFT and RL commits. Matching is deterministic, case-insensitive and on whole words, over the reasoning and visible text of the last three assistant messages up to and including the \texttt{final\_answer} call; rows show frequent matched phrases, not the full pattern list.}
\label{tab:app:markers}
\setlength{\tabcolsep}{4pt}
\begin{tabular}{@{}>{\raggedright\arraybackslash}p{0.29\linewidth}>{\raggedright\arraybackslash}p{0.68\linewidth}@{}}
\toprule
\textbf{Family} & \textbf{Frequent phrases} \\
\midrule
(1) Time or turn budget & ``given the time constraints'', ``time constraints'', ``running out of time'', ``turn budget'' \\
(2) Effort already spent & ``given all my research'', ``spent enough time'', ``given the extensive searching'', ``spent a lot of time'', ``given all the analysis'' \\
(3) Calls the answer a guess & ``best guess'', ``make an educated guess'', ``make my best guess'', ``make a guess'' \\
\bottomrule
\end{tabular}
\end{table}

\textbf{Method.} We compare the accuracy of flagged and unflagged commits within the same question, using questions where a model has both correct and wrong commits and both flagged and unflagged ones, weighting each question equally, with 95\% bootstrap intervals over questions. We then fit a regression with one intercept per question that also controls for whether the target (the gold answer, or the bridge entity for \stsql{}) had appeared in a tool result or in the agent's reasoning, and for rollout length.

\begin{figure}[t]
\centering
\includegraphics[width=0.8\linewidth]{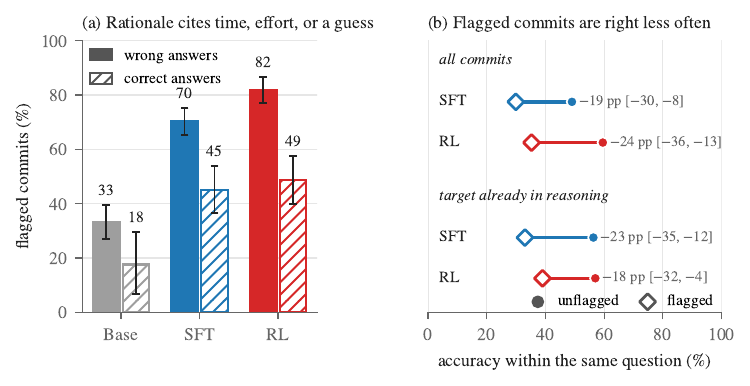}
\caption{A commit is flagged when its rationale (the last three assistant messages up to and including the \texttt{final\_answer} call) \textbf{(1) cites a time or turn budget}, \textbf{(2) cites the effort already spent}, or \textbf{(3) calls the answer a guess}. (a) Share of wrong (solid) and correct (hatched) commits that are flagged, pooled over rollouts; whiskers are 95\% bootstrap intervals over questions. (b) Accuracy of flagged (diamond) and unflagged (dot) commits within the same question (differences in percentage points, pp), for all commits and for commits where an assistant message up to the commit already named the target (the gold answer, or the bridge entity for \stsql{}).}
\label{fig:app:uncertain}
\end{figure}

\textbf{Results.} 82.0\% of RL's wrong commits are flagged against 48.7\% of its correct ones (SFT: 70.5\% vs 45.2\%; Figure~\ref{fig:app:uncertain}, Table~\ref{tab:app:uncertain}), and within the same question flagged commits are 24.3 points less often correct than unflagged ones for RL (SFT: 19.1); the gap stays negative after these controls (RL $-$10.0 [$-$16.2, $-$3.1], SFT $-$7.9 [$-$13.2, $-$1.8] points). The turn budget does not force the stop: the prompt calls it only ``limited'', the median SFT and RL rollouts end after 44 and 41 of 300 turns, and no SFT or RL rollout reaches the reminder sent with 10 turns left.

\begin{table}[!ht]
\centering\small
\caption{Supporting numbers for Figure~\ref{fig:app:uncertain}. (a) Coefficient on the flag in a linear probability model with one intercept per question over all commits, controlling for whether the target appeared in a tool result or in the agent's reasoning, and for log turns, context length and output tokens; the RL $-$ SFT difference is $-$2.1 [$-$11.0, +7.2]. (b) \sqlts{} and Parallel rollouts that keep working after the first assistant message containing a flag phrase: share correct at submission minus the share whose most recently named answer was correct at that message (observational). The target is the gold answer (for \stsql{}, the bridge entity); it is in the agent's reasoning if an assistant message up to the commit names it. The first two rows of the middle block use, for each model, the questions whose commits in that row are both correct and wrong and both flagged and unflagged (40--57 questions per estimate); intervals are 95\% bootstrap intervals over questions. The 120 questions include 12 verbatim-duplicate pairs; clustering them barely changes the estimates. Base submits an answer in only 49.2\% of rollouts, so its rates describe a selected subset, and the other rows are not estimated for it (--).}
\label{tab:app:uncertain}
\setlength{\tabcolsep}{4pt}
\begin{tabular}{@{}lccc@{}}
\toprule
 & \textbf{Base} & \textbf{SFT} & \textbf{RL} \\
\midrule
\multicolumn{4}{@{}l}{\emph{Flagged commits (\%)}} \\
\quad All commits & 30.7 & 64.3 & 72.9 \\
\quad Wrong commits & 33.2 & 70.5 & 82.0 \\
\quad Correct commits & 17.6 & 45.2 & 48.7 \\
\midrule
\multicolumn{4}{@{}l}{\emph{Accuracy of flagged minus unflagged commits, same question (points)}} \\
\quad All commits & -- & $-$19.1 {\scriptsize[$-$30.2, $-$7.7]} & $-$24.3 {\scriptsize[$-$36.0, $-$12.9]} \\
\quad Target already in the agent's reasoning & -- & $-$23.3 {\scriptsize[$-$34.9, $-$11.5]} & $-$18.0 {\scriptsize[$-$31.5, $-$4.1]} \\
\quad After controls for evidence and length (a) & -- & $-$7.9 {\scriptsize[$-$13.2, $-$1.8]} & $-$10.0 {\scriptsize[$-$16.2, $-$3.1]} \\
\midrule
\multicolumn{4}{@{}l}{\emph{Budget and continuation}} \\
\quad Rollouts that submit an answer (\%) & 49.2 & 96.8 & 100.0 \\
\quad Median turns per rollout (of 300) & 57 & 44 & 41 \\
\quad Rollouts reaching the 10-turns-left reminder & 15/960 & 0/960 & 0/960 \\
\quad Continuing past the first flag (points) (b) & -- & $-$3.7 {\scriptsize[$-$8.9, +0.8]} & $-$2.0 {\scriptsize[$-$6.5, +2.2]} \\
\bottomrule
\end{tabular}
\end{table}

\textbf{Conclusion.} Both trained agents often justify a submission by a time or turn budget, the effort already spent, or a guess, and such submissions are more often wrong; RL makes them more common (72.9\% vs 64.3\% of commits) with no detectable change in the gap after controls (RL $-$ SFT: $-$2.1 [$-$11.0, +7.2] points), and 28.6\% of the SFT training trajectories on hybrid questions, which were kept only when their answer was correct, already contain them. Since the turn budget does not stop the agent, a longer budget alone is unlikely to help; a promising direction is to make submission depend on evidence, for example through a verification step triggered by the flag.

\clearpage
\section{RL training and analysis}
\label{app:rl}

\subsection{RL training parameters}
\label{app:rl-recipe}

RL training is conducted using framework built on PRIME-RL \citep{primeintellect2025primerl}, with fully asynchronous training at asynchrony level 1, running on a single 8-GPU node. We optimize a GRPO objective~\citep{shao2024deepseekmath} with standard-deviation-normalized advantages and prompt-level token-averaged loss, using CISPO~\citep{minimaxm1} with importance weights clipped at $2.0$ and negative-sample scale $1.0$. We used AdamW~\citep{adamw} with learning rate $1\times10^{-6}$, $\epsilon=1\times10^{-15}$, $\beta_1=0.9$, $\beta_2=0.98$, zero weight decay, and a constant schedule without warmup. The reference-model KL penalty and explicit length penalty are disabled. Each step uses 32 prompts with group size 5. Rollouts use temperature $1.0$, top-$p=1.0$, up to 32k completion tokens per turn and max turns set to 100, with a 128k token context limit. Inference uses vLLM~\citep{vllm} with tensor parallelism 1 and 95\% GPU memory utilization. 

We additionally perform group-level difficulty filtering for adapting dynamic policy. Groups with 0\% or 100\% solve rates are excluded from the main optimization stream because they provide little within-group reward contrast. A task is removed from the training pool after being fully solved in $n=1$ consecutive rollout groups or failing completely in $m=3$ consecutive rollout groups. 

\subsection{Judge Voting Analysis}
\label{app:judge_examples}

As shown in Table~\ref{tab:app:judge_examples}, representative examples illustrate how five-vote majority judging resolves challenging cases involving semantic ambiguity and non-trivial variations in SQL result tables.

\begin{table*}[h]
\centering
\small
\caption{Representative cases where single-vote judging produces an incorrect reward while five-vote majority judging reaches the correct judgment. The SQL example illustrates that judging requires reasoning about result-level semantics rather than exact table matching.}
\label{tab:app:judge_examples}

\begin{tabular}{p{0.05\linewidth} p{0.88\linewidth}}
\toprule
\textbf{Case} & \textbf{Example} \\
\midrule

1 &
\begin{minipage}[t]{0.88\linewidth}
\textbf{Entity ambiguity.}
The reference entity is \texttt{Charles Emmanuel III}, while the model produces
\texttt{"Charles III"}. Although the two names are lexically similar, they refer
to different monarchs. A correct judgment therefore requires semantic entity
resolution rather than string matching.

\vspace{4pt}

\end{minipage}
\\[10pt]

2 &
\begin{minipage}[t]{0.88\linewidth}
\textbf{Partial SQL correctness.}
The gold result filters to \texttt{Fort de Pontisse}, while the model recovers
the same ranking but omits the final filter.

\vspace{4pt}
\begin{minipage}[t]{0.42\linewidth}
\centering
\textbf{Gold result}\\[2pt]
\begin{tabular}{lcc}
\toprule
\texttt{fort} & \texttt{avg\_days\_held} & \texttt{rank}\\
\midrule
Fort de Pontisse & 6.0 & 2\\
\bottomrule
\end{tabular}
\end{minipage}
\hfill
\begin{minipage}[t]{0.42\linewidth}
\centering
\textbf{Model result}\\[2pt]
\begin{tabular}{lcc}
\toprule
\texttt{fort} & \texttt{avg\_days} & \texttt{rank}\\
\midrule
Fort de Loncin   & 7.0 & 1\\
Fort de Pontisse & 6.0 & 2\\
Fort de Fléron   & 5.5 & 3\\
Fort de Flémalle & 3.5 & 4\\
Fort de Barchon  & 3.5 & 4\\
\bottomrule
\end{tabular}
\end{minipage}

\vspace{5pt}
The model correctly computes the averages and ranking, and correctly places
\texttt{Fort de Pontisse} at rank 2, but fails to apply the final filter.
The final result is therefore incorrect despite substantial partial correctness,
making the case difficult to judge reliably from surface-level table comparison.

\vspace{3pt}

\end{minipage}
\\

\bottomrule
\end{tabular}
\end{table*}

\end{document}